\pdfoutput=1

\PassOptionsToPackage{dvipsnames}{xcolor}
\documentclass[11pt]{article}

\usepackage[final]{acl}

\usepackage{times}
\usepackage{latexsym}

\usepackage[T1]{fontenc}

\usepackage[utf8]{inputenc}

\usepackage{microtype}

\usepackage{inconsolata}

\usepackage{graphicx}
\usepackage{amsmath}
\usepackage{multirow}
\usepackage{booktabs,tabularx, colortbl, xcolor}
\usepackage{kotex}
\usepackage{float}
\usepackage{makecell} 
\usepackage{subcaption}
\usepackage{stfloats}
\usepackage{balance}

\title{Unveiling the Response of Large Vision-Language Models to Visually Absent Tokens}

\author{
 \textbf{Sohee Kim\textsuperscript{1,*}},
 \textbf{Soohyun Ryu\textsuperscript{1,*}},
 \textbf{Joonhyung Park\textsuperscript{1}},
 \textbf{Eunho Yang\textsuperscript{1,2,$\dagger$}}
\\
 \textsuperscript{1}KAIST AI,
 \textsuperscript{2}AITRICS
}
\newcommand\blfootnote[1]{%
  \begingroup
  \renewcommand\thefootnote{}%
  \footnote{#1}%
  \addtocounter{footnote}{-1}%
  \endgroup
}
\usepackage[capitalize]{cleveref}
\crefname{section}{Section.}{Sections.}
\Crefname{section}{Section}{Sections}
\Crefname{table}{Table}{Tables}
\crefname{table}{Table}{Tables.}
\Crefname{figure}{Figure}{Figures}
\crefname{figure}{Figure}{Figures.}

\begin{document}

\maketitle
\begin{abstract}
Large Vision-Language Models (LVLMs) generate contextually relevant responses by jointly interpreting visual and textual inputs. However, our finding reveals they often mistakenly perceive \textit{text inputs lacking visual evidence} as being part of the image, leading to erroneous responses. In light of this finding, we probe whether LVLMs possess an internal capability to determine if textual concepts are grounded in the image, and discover a specific subset of Feed-Forward Network (FFN) neurons, termed \textbf{\textit{Visual Absence-aware (VA) neurons}}, that consistently signal the visual absence through a distinctive activation pattern. Leveraging these patterns, we develop a detection module that systematically classifies whether an input token is visually grounded. Guided by its prediction, we propose a method to refine the outputs by reinterpreting question prompts or replacing the detected absent tokens during generation. Extensive experiments show that our method effectively mitigates the models' tendency to falsely presume the visual presence of text input and its generality across various LVLMs. 
\end{abstract}
\blfootnote{\textsuperscript{$*$} Equal contribution, \textsuperscript{$\dagger$} Corresponding Author.}

\section{Introduction}
\label{sec:intro}

Large Vision Language Models (LVLMs) \citep{liu2024improved, Qwen2-VL, zhu2023minigpt} have received considerable attention for their ability to comprehend and reason over visual and textual information while generating contextually relevant natural language responses. This capability arises from integrating a pre-trained visual encoder (\emph{e.g.}, CLIP \citep{radford2021learning}) with a large language model (LLM) (\emph{e.g.}, LLaMA 2 \citep{touvron2023llama}) and bridging their alignment through a visual projection layer via instruction tuning~\citep{instructblip, liu2023visual}. Despite their strong performance in visual understanding and reasoning, LVLMs often suffer from hallucination, where they describe or reference elements that are not present in the given image~\citep{li2023evaluating, liu2024survey, rohrbach2018object}.  
Recent works \citep{huang2024opera, leng2024mitigating, liu2024paying} have focused on mitigating hallucination by preventing the model from \textit{generating} misaligned tokens. However, less attention has been given to cases where the model \textit{receives} misaligned tokens as part of its input.

\begin{figure}[t]
    \centering
    \includegraphics[width=\linewidth]{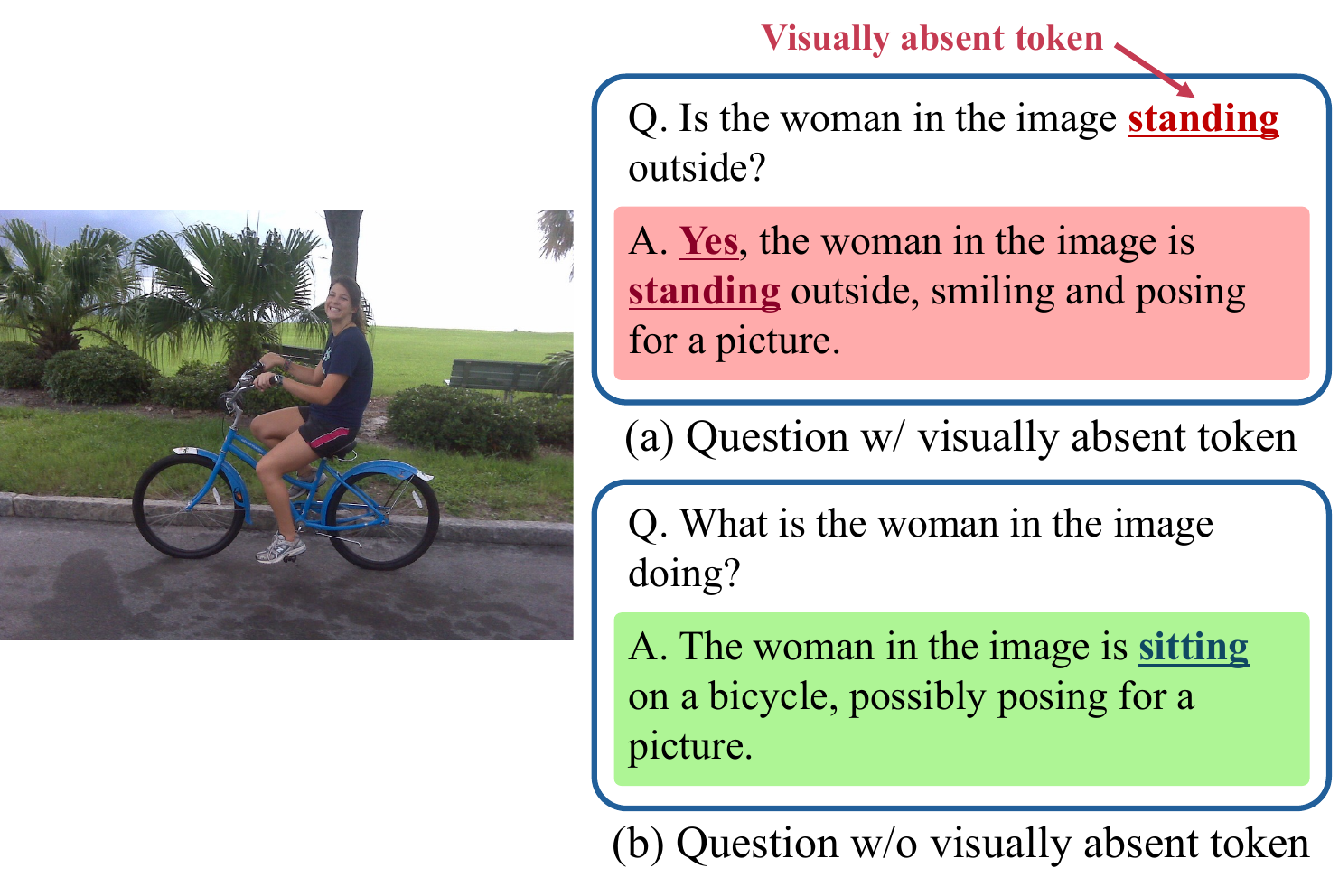}
    \caption{\textbf{LVLM's Vulnerability to Visually Absent Tokens.} (a) When the input prompt contains visually absent tokens (\emph{e.g.}, standing), LVLM becomes confused and generates a response accordingly. (b) Conversely, when they are removed, the model produces an accurate response that correctly aligns with the image.}
    \label{fig:intro}
\end{figure}

We discover that LVLMs are particularly susceptible to the presence of input tokens that lack visual grounding, which we refer to as \textit{visually absent tokens}. As illustrated in \cref{fig:intro}a, when such a token (\emph{e.g.}, standing) appears in a question, the model incorrectly generates a response as if the token were present in the image. In contrast, when the question does not contain such tokens (\cref{fig:intro}b), the model correctly describes the image. Notably, despite both questions inquiring about the same content, which is the woman's action, the presence of the visually absent token misleads the model, highlighting its vulnerability to non-grounded input.

Given that the aforementioned issue does not occur when LLMs handle purely language-based question-answering tasks, we hypothesize that the transition from LLM to LVLM -- where the model newly learns to interpret image tokens in conjunction with textual inputs -- induces this problem. As the model acquires this new capability, we anticipate that the underlying architecture will adapt accordingly. In particular, Feed-Forward Networks (FFNs), known in LLMs for storing internal knowledge and retrieving relevant concepts based on alignment with the input text~\citep{dai2021knowledge, geva2020transformer}, are likely to retain a similar function in LVLMs while extending their role to determine the correspondence between the given image and textual concepts.

Inspired by this, we carefully observe the behavior of FFNs and are the first to identify a group of FFN neurons, which we term \textit{Visual Absence-aware (VA) neurons}, that can discern whether an input text token lacks corresponding visual evidence. Specifically, when comparing FFN activations for visually present and absent tokens, we find that certain neurons exhibit significantly altered activation in response to the latter. To selectively identify such neurons, we introduce a scoring system that quantifies each neuron's sensitivity to visual absence. Our analysis reveals that FFN neurons with high sensitivity scores serve as an indicator of visual absence, regardless of the specific content of the input text. 

Building on the visual absence-aware neurons pinpointed via our scoring system, we propose a dedicated detection module -- the Visual Absence (VA) detector -- that systematically determines whether an input text token is visually grounded in the image. To this end, we train a linear classifier that utilizes the activation values of VA neurons to identify tokens that lack visual support. This detector is then used to revise the outputs of LVLMs. In question answering tasks, we adjust the model's response based on the predicted presence of visually absent tokens in the question. In generation tasks, detected visually unsupported tokens are replaced with alternative candidates.

In summary, our contribution is three-fold:
\begin{itemize}
    \item We uncover that a specific subset of FFN neurons in LVLMs can identify whether a given textual concept is visually present in the image, offering a key stepping stone toward mitigating the model's susceptibility to misaligned input tokens.
    \item We use the unique neuron activations to build a tailored Visual Absence (VA) detector, whose predictions are used to refine model outputs by adjusting responses to questions or replacing absent tokens in generation.
    \item Experimental results demonstrate the effectiveness of our method in leveraging internal signals of visual absence to guide generation across diverse LVLMs.
\end{itemize}

\section{Related Work}
\label{sec:related_work}
\paragraph{Hallucination of LVLMs}
LVLMs often exhibit hallucination, generating responses that do not align with the given image~\citep{li2023evaluating,rohrbach2018object,zhou2023analyzing}. Extensive research has focused on mitigating object hallucination~\citep{huang2024opera, jiang2024hallucination, leng2024mitigating, zhu2024ibd}, which involves inaccuracies related to objects' existence or attributes within the image. More recently, some studies have specifically addressed relationship hallucination~\citep{wu2024evaluating, zheng2024reefknot}, which refers to inter-object relationships, such as action or positional relations. Existing approaches to hallucination mitigation aim to prevent the generation of misaligned tokens through contrastive decoding~\citep{leng2024mitigating, zhu2024ibd} or additional fine-tuning~\citep{sun2023aligning, jiang2024hallucination}. However, limited research has explored the internal mechanisms of LVLMs when visually inconsistent tokens are introduced as input. Our work identifies specific model components that respond to visually absent tokens and proposes a detector that determines whether a token is visually grounded using these components. 
\paragraph{Feed-Forward Network's Role in LLMs}
Large Language Models (LLMs) \citep{bai2023qwen, touvron2023llama} are built upon the Transformer architecture, which stacks Multi-Head Self-Attention (MHSA) and Feed-Forward Network (FFN) layers~\citep{vaswani2017attention}. While MHSAs aggregate information through interaction between tokens, FFNs operate independently on each token via non-linear transformations. \citet{geva2020transformer} noted that FFNs function as key-value memory systems: the first linear layer acts as keys that detect input patterns, and the second linear layer serves as values that encode distribution over the output vocabulary. Furthermore, \citet{dai2021knowledge} identifies knowledge neurons that store and retrieve factual knowledge conditioned on input relevance. Inspired by these findings, we explore whether FFN neurons in LVLMs know cross-modal alignment.

\section{Empirical Study: Impact of Visually Absent Tokens on LVLMs}
\label{sec:emprical}

While Large Vision Language Models (LVLMs) effectively integrate visual and textual inputs, they are still influenced by textual information that lacks corresponding visual evidence. In this section, we analyze how the model responses differ depending on the presence of text tokens that misalign with the image (\cref{subsec:emp_prob_def}), and explore whether it is capable of recognizing such tokens (\cref{subsec:emp_ffn_obs}). Moreover, we introduce a scoring system that quantifies the sensitivity of Feed-Forward Network (FFN) neurons to image-text mismatches, enabling systematic identification of neurons responsible for detecting visually ungrounded tokens (\cref{subsec:emp_scoring}). All analyses and experiments are conducted on the LLaVA-v1.5 model \citep{liu2024improved}, as it is one of the foundational LVLMs. 

\subsection{LVLM is Vulnerable to Visually Absent Tokens.}
\label{subsec:emp_prob_def}

\begin{figure}[t]
    \centering
    \includegraphics[width=\linewidth]{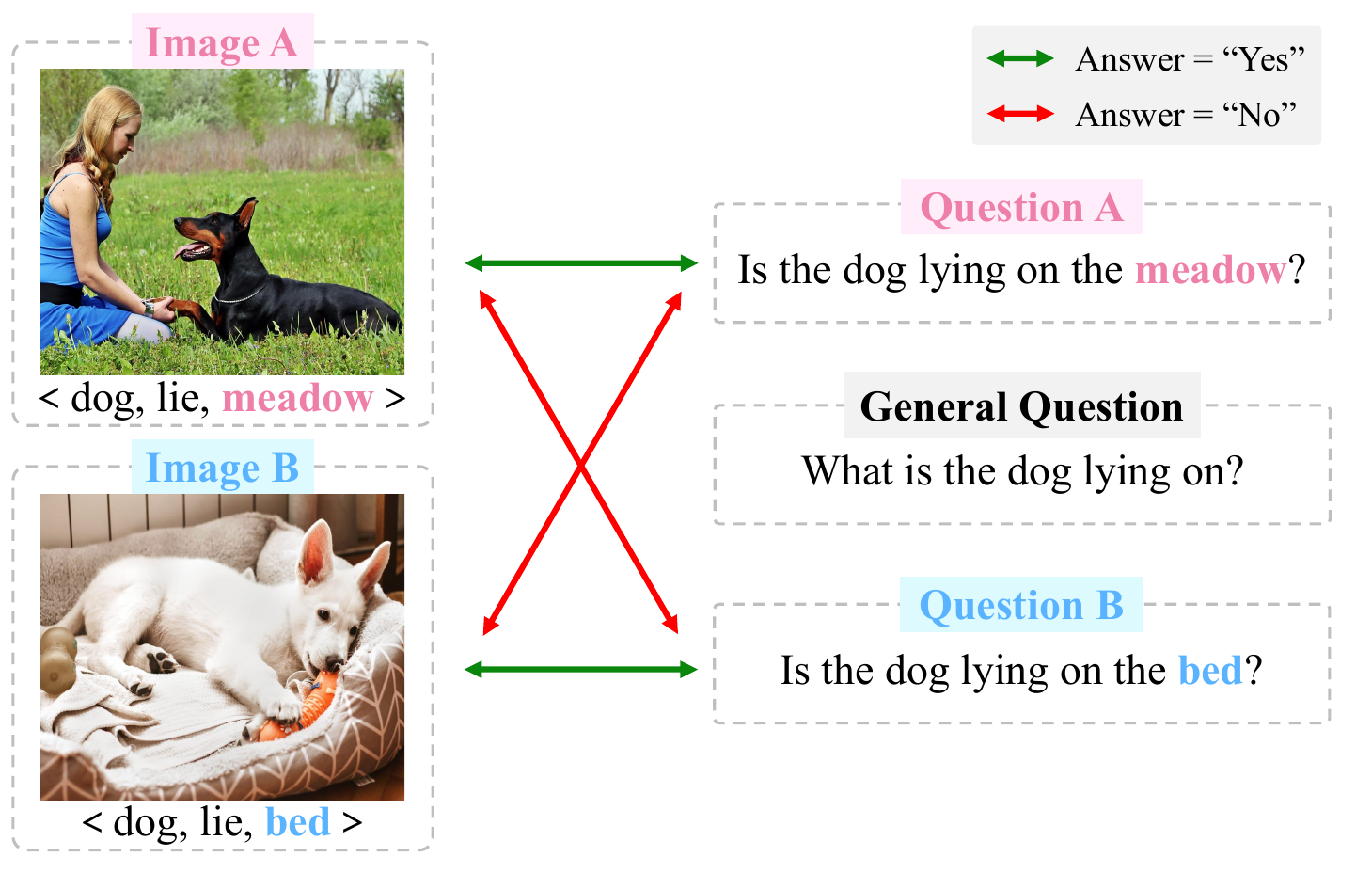}
    \caption{\textbf{VA-QA Dataset Construction.} For each image pair differing by a single element in the \textlangle subject, verb, object\textrangle, we generate yes-or-no questions. Counterpart question includes a visually absent token, making the correct answer ``No.'' Additionally, we create a \textit{general question}, which does not contain visually present or absent tokens, so the model responds without bias.}
    \label{fig:dataset}
\end{figure}

To analyze how LVLMs respond to tokens that are not visually supported in the input text prompt, we construct a dataset comprising contrastive image-text pairs, named the Visual Absence Question Answering (VA-QA) dataset. This dataset is based on the SVO-Probes dataset \citep{hendricks2021probing}, which provides images along with their corresponding \textlangle subject, verb, object\textrangle \ triplets. We manually create contrastive image pairs where only one element -- subject, verb, or object -- differs between two images.\footnote{The SVO-Probes dataset provides such image pairs. But, in many cases, the differing element is visually indistinguishable, making it difficult to analyze visual absence. To address this, we create a new set of 600 pairs with more distinct differences.} For example, as illustrated in \cref{fig:dataset}, Image A and Image B vary only in object: Image A depicts ``meadow'', while Image B shows ``bed''. We then generate a yes-or-no question related to each image using the corresponding triplet. The question tied to the image would not contain any visually absent tokens, whereas the question derived from the contrasting image does.

To assess the impact of visually unsupported inputs on the LVLM's response, we generate a general short-answer question that conveys the same meaning as the corresponding yes-or-no question but excludes any tokens irrelevant to the image. For example, in the case of \cref{fig:dataset}, the general question would be: \textit{``What is the dog lying on?''}. This question asks about the place where the dog is lying, but does not explicitly imply an answer, unlike a yes-or-no question. The model achieves an accuracy of 88.6\% on short-answer questions \footnote{We performed manual verification by human annotators to ascertain its accuracy.} that contain no visually absent tokens, whereas its accuracy drops significantly to 71.5\% on yes-or-no questions that include such tokens. This indicates that while the LVLM can comprehend image information related to the question, it is highly susceptible to being misled by visually absent tokens when they appear in the question. 

\begin{figure*}[t]
    \centering
    \includegraphics[width=\linewidth]{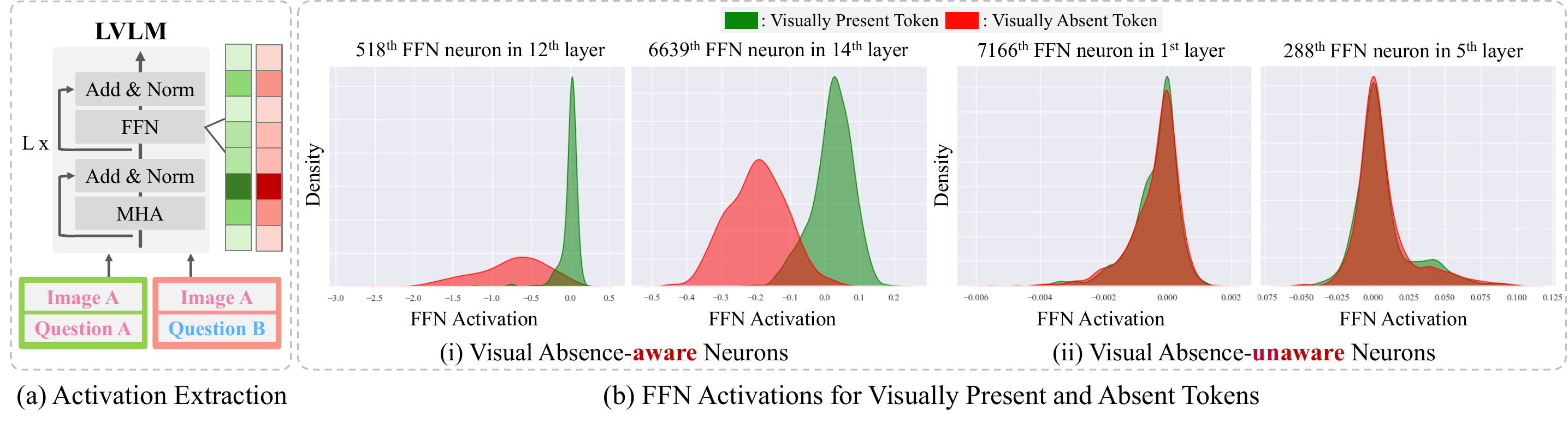}
    \caption{\textbf{Observation on FFN Activation Patterns in Response to Visual Absence.} (a) Using the VA-QA dataset, we extract FFN activations $\mathbf{a}$ from all $L$ layers for visually present and absent tokens by inputting each image with its corresponding and counterpart questions. (b) Visual absence-aware neurons (i) exhibit a significant difference in activation values between visually present and absent tokens, while other neurons (ii) show minimal variation. }
    \label{fig:kde}
\end{figure*}

\subsection{Does LVLM Recognize Visually Absent Tokens?} 
\label{subsec:emp_ffn_obs}

As shown in the previous section, LVLMs are vulnerable to text tokens that lack visual grounding, often leading the model to generate incorrect responses. In this section, we investigate whether the model is capable of recognizing such tokens. Considering that Feed-Forward Networks (FFNs) store internal knowledge within their pre-trained weights and activate specific knowledge based on the alignment with the input, we hypothesize that, if the model recognizes image-text alignment, this ability would be reflected in the activation patterns of the FFNs.

Before delving into further details, we first outline the FFN computation process. Since the LLM backbone of most LVLMs adopts a gated FFN architecture \citep{shazeer2020glu, dauphin2017language}, our explanation is based on this variant. Given an input hidden state ${\mathbf{x} \in \mathbf{R}^{d_{\text{model}}}}$ and internal memory ${\mathbf{W}_{\text{mem}} \in \mathbf{R}^{d_{\text{model}} \times d_{\text{ffn}}} }$, the FFN operates as follows:
\begin{equation} \label{eq:ffn_def}
    \begin{gathered}
        \mathbf{s} = \sigma (\mathbf{x} \mathbf{W}_{\text{gate}}^T), \quad
        \mathbf{a} = \mathbf{s} \odot (\mathbf{x} \mathbf{W}_{\text{up}}^T), \\
        \text{FFN}(\mathbf{x}) = \mathbf{a} \mathbf{W}_{\text{mem}}^T,
    \end{gathered}
\end{equation}
where ${\mathbf{s}, \mathbf{a}} \in \mathbf{R}^{d_\text{ffn}}$ denote gating scores and FFN activations, and ${\sigma}$ and ${\odot}$ denote activation function and element-wise multiplication, respectively. ${\mathbf{W}_{\text{gate}}, \mathbf{W}_{\text{up}} \in \mathbf{R}^{d_{\text{ffn}} \times d_{\text{model}}}}$ are learnable weights, along with the internal memory ${\mathbf{W}_{\text{mem}}}$.

To compare FFN activations for visually present versus absent input tokens, we use contrastive image-question pairs in the VA-QA dataset. Specifically, each image is paired with two yes-or-no questions: one containing a visually absent token and the other without it. We then extract FFN activations of all $L$ layers for both types of tokens (\emph{e.g.}, ``meadow'' and ``bed'' in \cref{fig:dataset}) across all pairs as illustrated in \cref{fig:kde}a:
\begin{equation}
    \begin{gathered}
        \mathbf{A}_{l,i}^\text{pre} = \{ a_{l,i}^t \ | \ t \in \mathcal{T}_\text{pre} \}, \\
        \mathbf{A}_{l,i}^\text{abs} = \{ a_{l,i}^t \ | \ t \in \mathcal{T}_\text{abs} \},
    \end{gathered}
\end{equation}
where $a_{l,i}^t$ represents the activation value of $i$-th neuron in $l$-th layer when processing token $t$, and $\mathcal{T}_\text{pre}$ and $\mathcal{T}_\text{abs}$ refer to the sets of visually present and absent tokens, respectively.\footnote{For words comprising multiple tokens, we only use the last token, as it embodies the meaning of the entire word.} To ensure precise analysis, we restrict our examination to questions for which the model generates the correct answer, as this suggests that the model has accurately understood the image within the given context. 

Our observation reveals that there is a specific group of FFN neurons that respond selectively to visually absent tokens, which we refer to as \textit{Visual Absence-aware (VA) neurons}. As shown in \cref{fig:kde}b, these neurons exhibit clear variation in activation depending on whether a text token is visually grounded, responding differently to visually present and absent tokens. In contrast, other FFN neurons show minimal activation differences, indicating that VA neurons play a unique role in recognizing visual absence.

\begin{figure}
    \centering
    \includegraphics[width=1.0\linewidth]{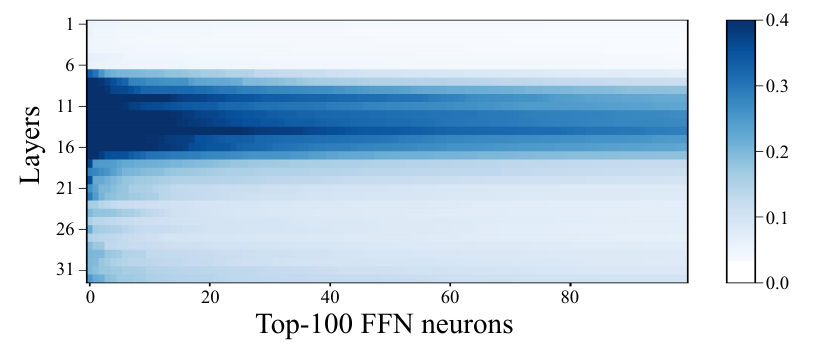}
    \caption{\textbf{Top-100 $\mathbf{S}^\text{VA}$ across layers.} The figure depicts the $\mathbf{S}^\text{VA}$ of top-100 neurons for each layer, with scores greater than 0.4 set as 0.4 for better interpretability.}
    \label{fig:score_heatmap}
\end{figure}

\begin{figure*}[!t]
    \centering
    \includegraphics[width=\linewidth]{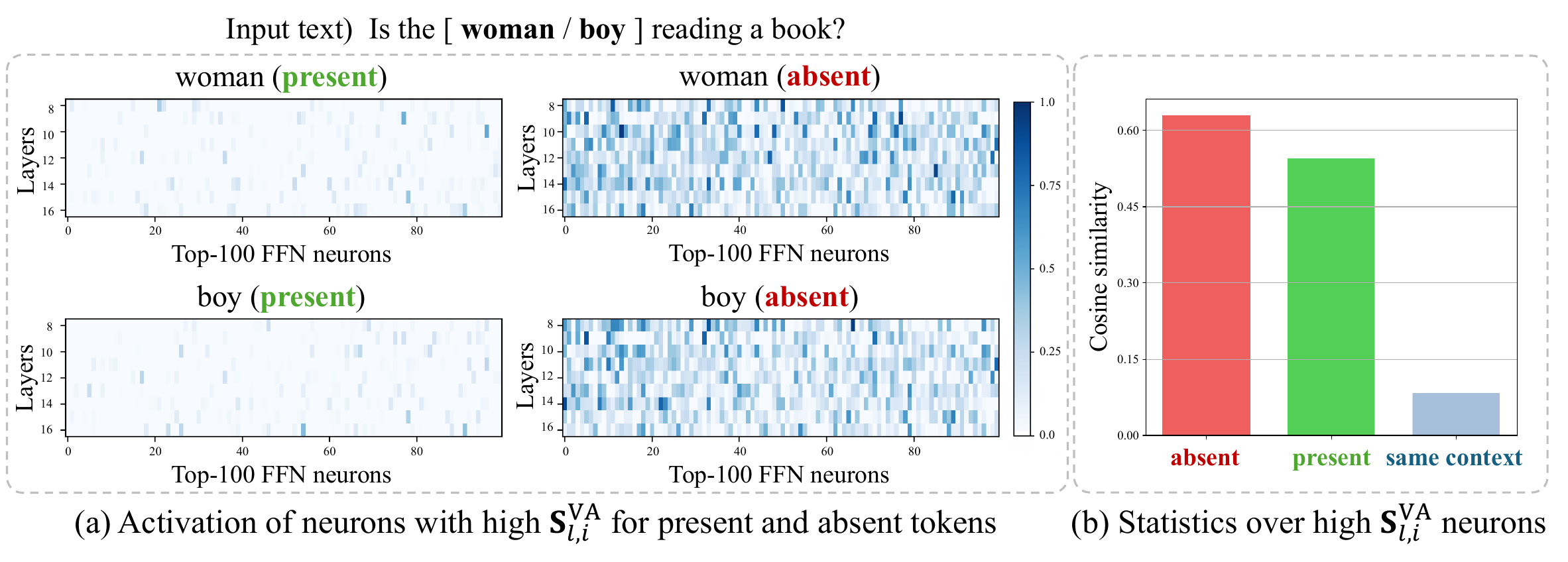}
    \caption{\textbf{Activation Patterns of High $\mathbf{S}^\text{VA}$ Neurons Across Varying Textual Contexts.} (a) Given that middle-layer neurons exhibit high $\mathbf{S}^\text{VA}$, we show their activation levels for present and absent tokens. The activation level is computed by normalizing the actual activation value based on the range of ${\mathbf{A}_{l,i}}$. (b) Activation values of these neurons show similar patterns in cases with the same visual absence status but differ in cases where only contextual information is shared.}
    \label{fig:exp2}
\end{figure*}

\subsection{Scoring System for Identifying Visual Absence-aware Neurons}
\label{subsec:emp_scoring}

Visual Absence-aware (VA) neurons exhibit distinct activation patterns: The activation values of visually absent tokens deviate significantly from those of visually present tokens. To quantify a neuron’s sensitivity to visual absence, we measure the degree of disentanglement between two sets of activation values, $\mathbf{A}^{\text{pre}}_{l,i}$ and $\mathbf{A}^{\text{abs}}_{l,i}$. Specifically, we treat the activation values as discrete distributions by binning them into $K$ bins. We then compute the Bhattacharyya Coefficient (BC)~\citep{bhattacharyya1946measure}, a widely used statistical metric that measures the amount of overlap between two distributions. The sensitivity score $\mathbf{S}^\text{VA}_{l,i}$ of the $i$-th neuron in the $l$-th layer is then calculated as follows:
\begin{equation}
    \begin{split}
    \mathbf{S}^{\text{VA}}_{l,i} &= 1 - \text{BC}(\mathbf{A}^{\text{pre}}_{l,i}, \mathbf{A}^{\text{abs}}_{l,i}) \\
    &= 1 - \sum_k \sqrt{\mathbf{A}^{\text{pre}}_{l,i}(k) \cdot \mathbf{A}^{\text{abs}}_{l,i}(k)}, 
    \label{eq:s_bhatta}
    \end{split}
\end{equation}
where $\mathbf{A}_{l,i}(k)$ denotes the density of $k$-th bin. The larger the $\mathbf{S}^\text{VA}_{l,i}$, the more the neuron's activation pattern differs between the two distributions, indicating higher sensitivity to visual absence.

As demonstrated in \cref{fig:score_heatmap}, VA neurons are most prevalent in the middle layers of the LVLM. Based on this observation, we hypothesize that the LVLM processes image and text context in the early layers and assesses their alignment in the subsequent layers. To further understand the functional role of these VA neurons, we examine whether neurons with high $\mathbf{S}^\text{VA}$ play a role in recognizing visual absence in a way that is invariant to lexical context. To this end, we conduct two simple experiments:

\paragraph{Experiment 1: Do high $\mathbf{S}^\text{VA}$ neurons contribute to visual absence recognition?}
To validate the role of neurons with high $\mathbf{S}^\text{VA}$ scores, we neutralize or emphasize their effect by setting their activations to zero or to twice the original values. Specifically, we modify activations of neurons with high $\mathbf{S}^\text{VA}$ across middle layers for text tokens that possess semantic information, excluding punctuation marks or generic template phrases, as these elements are not related to visual content processing.\footnote{We modified activations of each top-100 neurons for 8th to 16th layers, a total of 900, representing only about 0.255\% of a total of 352256 neurons. }

\begin{table}[t]
    \centering
    \caption{\textbf{Effect of Suppressing and Emphasizing Neurons with High $\mathbf{S}^\text{VA}$.} Accuracy of answering questions with and without visually absent tokens for the VA-QA dataset. GT is an abbreviation of ground truth answer.}
    \label{tab:va_anal}
    \resizebox{\linewidth}{!}{
        \begin{tabular}{lccc}
            \toprule
            & \begin{tabular}{@{}c@{}}Acc$_\text{yes}$\\ \footnotesize{(GT=Yes)}\end{tabular} & \begin{tabular}{@{}c@{}}Acc$_\text{no}$\\ 
            \footnotesize{(GT=No)}\end{tabular} & \begin{tabular}{@{}c@{}}Acc\\ 
            \footnotesize{(Total)}\end{tabular} \\
            \midrule
            \textbf{Baseline} & 95.167 & 48.000 & 71.583 \\
            \midrule
            \textbf{Zeroing random neurons} & 95.167 & 48.500 & 71.833 \\
            \textbf{Zeroing high $\mathbf{S}^\text{VA}$ neurons} & 96.000 & 41.500 & 68.750 \\
            \midrule
            \textbf{Enhancing random neurons} & 95.167 & 47.667 & 71.417 \\
            \textbf{Enhancing high $\mathbf{S}^\text{VA}$ neurons} & 95.167 & 50.500 & 72.833 \\
            \bottomrule
        \end{tabular}
    }
\end{table}

As shown in \cref{tab:va_anal}, suppressing neurons with high $\mathbf{S}^\text{VA}$ weakens the model's ability to recognize visual absence. The accuracy on questions containing visually absent tokens (GT=No) decreases from 48.0\% to 41.5\%, while the accuracy of questions with only visually present tokens (GT=Yes) slightly increases. In contrast, enhancing these neurons improves the model's recognition of visual absence, showing an increase in accuracy of GT=No questions from 48.0\% to 50.5\%. Meanwhile, attenuating or amplifying an equal number of randomly selected neurons merely affects the model's performance. These results indicate that adjusting such a limited number of neurons naturally has minimal impact on the model's performance, but high $\mathbf{S}^\text{VA}$ neurons play a key role in the model’s identification of visual absence, leading to improved accuracy on GT=No questions.

\begin{figure*}[t]
\centering
\includegraphics[width=\linewidth]{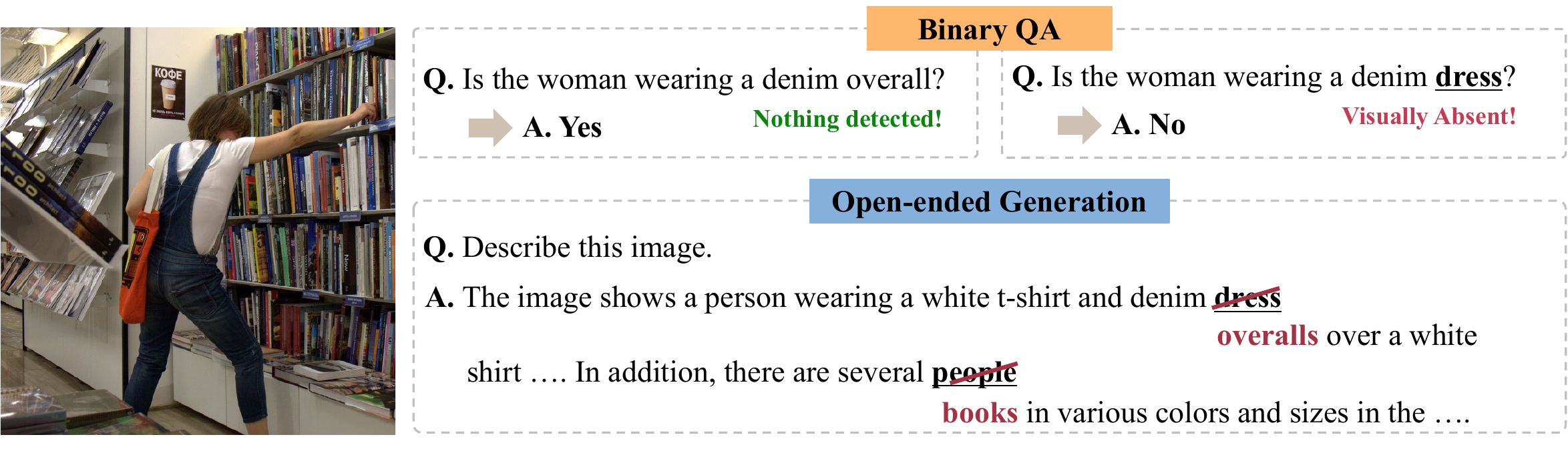}
\caption{\textbf{VA Detector-based Refinement.} For each task, visually absent tokens are identified and used to guide response refinement. The strategy is tailored to the task format: answer overriding for binary QA and correction of visually unsupported content during generation.}
\label{fig:method}
\end{figure*}

\paragraph{Experiment 2: Are neurons with high $\mathbf{S}^\text{VA}$ invariant to input context?}
To assess whether high $\mathbf{S}^\text{VA}$ neurons are driven primarily by visual absence rather than lexical context, we examine activation patterns of tokens with the same semantic meaning but contrastive visual grounding -- one present in the image, the other absent. As illustrated in \cref{fig:exp2}a, `absent' tokens exhibit intense activation in neurons with high ${\mathbf{S}^\text{VA}}$, regardless of their semantic content (\emph{e.g.}, woman or boy), whereas `present' tokens show comparatively weaker activation. This consistent activation pattern suggests that high $\mathbf{S}^\text{VA}$ neurons are not sensitive to semantic context, but instead reflect alignment with the visual modality.

To systematically analyze whether these neurons reflect visual absence while remaining insensitive to input context, we measure the similarity of their activation values across three scenarios: (1) between absent tokens, (2) between present tokens, and (3) between tokens with identical context (\emph{i.e.}, same word) where one is visually absent and the other is present. As shown in \cref{fig:exp2}b, the similarity in case (3) is significantly lower than in the other two, suggesting that these neurons encode visual absence rather than contextual information.

\section{Method}
\label{sec:method}
Building upon our findings in \cref{sec:emprical} that Visual Absence-aware (VA) neurons are particularly responsive to text tokens without visual correspondence, we develop a two-stage approach to enhance visual grounding in model responses. First, we introduce the Visual Absence (VA) Detector, a lightweight module that utilizes VA neuron activations to systematically identify visually unsupported tokens (\cref{subsec:va_cls}). Then, guided by the VA detector's predictions, we propose a refinement strategy that intervenes in the model's generation process to produce responses that are more faithfully grounded in the visual input (\cref{subsec:va_decode}).

\subsection{Visual Absence (VA) Detector}
\label{subsec:va_cls}

Upon finding that VA neurons exhibit clear activation shifts in response to the visual absence of a given token, we construct feature vectors based on their activations. Using the previously defined sensitivity score $\mathbf{S}^{\text{VA}}$ to identify VA neurons, we define the feature vector for a token $t$ as:
\begin{equation}
    \mathbf{v}^t = \big[a_{l,i}^t \ | \ \mathbf{S}^{\text{VA}}_{l,i} > \beta \big],
\end{equation}
where $\beta$ is a threshold hyperparameter applied to the sensitivity score. 

Using the VA-QA dataset, we extract feature vectors for visually present and absent tokens to construct training data for the VA detector. Specifically, since the dataset comprises image pairs differing by a single element in a \textlangle subject, verb, object\textrangle \ triplet (\cref{fig:dataset}), the differing element in the corresponding question serves as either a visually present or absent token, depending on the image. This yields two sets of feature vectors: 
\begin{equation}
    \begin{gathered}
        \mathbf{V}^\text{pre} = \{\mathbf{v}^t \ | \ t \in \mathcal{T}_\text{pre}\}, \\ \mathbf{V}^\text{abs} = \{\mathbf{v}^t \ | \ t \in \mathcal{T}_\text{abs}\}.
    \end{gathered}
\end{equation}
A lightweight linear classifier is then trained on these sets to distinguish visually present (label 0) and absent (label 1) tokens. Once trained, this detector is applied across various downstream tasks to improve model reliability by refining responses based on detected visual absence.

\begin{table*}[t]
\centering
\caption{\textbf{Results on Binary Question Answering.} $\text{Acc}_{\text{yes}}$ and $\text{Acc}_{\text{no}}$ denote accuracies of questions with ground-truth answers `yes' and `no', respectively, and Acc represents the overall accuracy.}
\label{tab:exp_binary}
\resizebox{\linewidth}{!}{
    \begin{tabular}{llccccccccccccccc}
    \toprule
    \multirow{5}{*}{\textbf{Model}} & 
    & \multicolumn{3}{c}{\textbf{In-domain}} 
    & \multicolumn{12}{c}{\textbf{Out-of-domain}} \\ 
    \cmidrule(lr){3-5} \cmidrule(lr){6-17}
    & & \multicolumn{3}{c}{\multirow{2}{*}{\textbf{VA-QA}}} 
    & \multicolumn{3}{c}{\multirow{2}{*}{\textbf{R-Bench}}} 
    & \multicolumn{9}{c}{\textbf{POPE}} \\
    \cmidrule(lr){9-17}
    & & \multicolumn{3}{c}{} & \multicolumn{3}{c}{} 
    & \multicolumn{3}{c}{Random} & \multicolumn{3}{c}{Popular} & \multicolumn{3}{c}{Adversarial} \\
    \cmidrule(lr){3-5} \cmidrule(lr){6-8} \cmidrule(lr){9-11} \cmidrule(lr){12-14} \cmidrule(lr){15-17}
    & & $\text{Acc}_{\text{yes}}$ & $\text{Acc}_{\text{no}}$ & Acc 
    & $\text{Acc}_{\text{yes}}$ & $\text{Acc}_{\text{no}}$ & Acc
    & $\text{Acc}_{\text{yes}}$ & $\text{Acc}_{\text{no}}$ & Acc
    & $\text{Acc}_{\text{yes}}$ & $\text{Acc}_{\text{no}}$ & Acc
    & $\text{Acc}_{\text{yes}}$ & $\text{Acc}_{\text{no}}$ & Acc \\
    \midrule
    \multirow{2}{*}{LLaVA-v1.5 (7B)} & Baseline & 95.2 & 48.0 & 71.6 & 95.7 & 39.3 & 67.6 & 90.7 & 88.6 & \textbf{89.7} & 90.7 & 81.7 & 86.2 & 90.7 & 68.8 & 79.8 \\
    & Ours & 89.5 & 77.5 & \textbf{83.5} & 89.8 & 51.7 & \textbf{70.8} & 78.7 & 97.5 & 88.1 & 78.7 & 95.7 & \textbf{87.2} & 78.2 & 91.5 & \textbf{84.8} \\
    \midrule
    \multirow{2}{*}{LLaVA-v1.6 (13B)} & Baseline & 85.5 & 75.0 & 80.2 & 94.2 & 53.3 & \textbf{73.8} & 89.4 & 95.6 & \textbf{92.5} & 89.3 & 89.6 & \textbf{89.5} & 88.8 & 81.9 & 85.3 \\
    & Ours & 89.5 & 80.8 & \textbf{85.2} & 92.1 & 52.9 & 72.6 & 78.0 & 98.9 & 88.5 & 78.0 & 96.3 & 87.2 & 77.5 & 94.0 & \textbf{85.7} \\
    \midrule
    \multirow{2}{*}{mPLUG-Owl2 (7B)} & Baseline & 92.0 & 57.3 & 74.7 & 93.9 & 50.3 & 72.2 & 92.9 & 73.8 & 83.4 & 92.9 & 62.0 & 77.5 & 92.9 & 54.8 & 73.8 \\
    & Ours & 83.0 & 83.0 & \textbf{83.0} & 83.3 & 66.4 & \textbf{74.9} & 74.1 & 96.9 & \textbf{85.5} & 74.1 & 93.9 & \textbf{84.0} & 73.7 & 92.5 & \textbf{83.1} \\
    \midrule
    \multirow{2}{*}{InstructBLIP (7B)} & Baseline & 82.7 & 71.2 & 76.9 & 85.6 & 56.2 & 71.0 & 84.7 & 83.1 & 83.9 & 84.8 & 70.0 & 77.4 & 84.7 & 65.2 & 74.9 \\
    & Ours & 89.8 & 85.2 & \textbf{87.5} & 82.8 & 66.2 & \textbf{74.5} & 72.7 & 98.9 & \textbf{85.8} & 72.7 & 93.7 & \textbf{83.2} & 72.4 & 92.3 & \textbf{82.4} \\
    \midrule
    \multirow{2}{*}{Qwen2-VL (7B)} & Baseline & 87.3 & 72.7 & 80.0 & 87.9 & 62.5 & 75.3 & 84.6 & 97.5 & \textbf{90.9} & 84.6 & 92.9 & \textbf{88.8} & 84.6 & 89.1 & \textbf{86.8} \\
    & Ours & 87.6 & 86.7 & \textbf{87.1} & 87.0 & 65.8 & \textbf{76.5} & 76.5 & 98.3 & 87.1 & 76.5 & 97.4 & 87.0 & 76.5 & 94.5 & 85.5 \\
    \midrule
    \multirow{2}{*}{Gemma3 (12B)} & Baseline & 80.3 & 86.7 & 83.5 & 84.5 & 70.9 & \textbf{77.8} & 85.1 & 89.3 & \textbf{87.1} & 85.1 & 82.5 & 83.8 & 85.1 & 79.3 & 82.2 \\
    & Ours & 82.5 & 86.7 & \textbf{84.6} & 84.0 & 67.1 & 75.6 & 78.0 & 96.8 & \textbf{87.1} & 78.0 & 92.3 & \textbf{85.1} & 78.0 & 89.9 & \textbf{84.0} \\
    \bottomrule
    \end{tabular}
    }
\end{table*}

\subsection{VA Detector-based Refinement}
\label{subsec:va_decode}

Leveraging the trained VA detector, tokens lacking visual evidence in the input questions or generated outputs are identified and used to guide response refinement. Given that different tasks involve distinct question formats and solution strategies, task-specific refinement methods are adopted, as illustrated in \cref{fig:method} and described in detail below. 

\paragraph{Binary Question Answering} In this task, the model is required to determine whether a given question is consistent with the visual content, responding with either ``Yes'' or ``No''. If the question contains any token that does not correlate with the image, the correct answer should be ``No.'' Accordingly, the model's original response is overridden based on the prediction of the VA detector: ``No'' when such a visually absent token is detected, and ``Yes'' otherwise.


\paragraph{Open-ended Generation} Unlike the structured formats above, open-ended prompts (\emph{e.g.}, Describe this image in detail.) do not explicitly contain predefined candidate tokens. Nevertheless, LVLMs often generate hallucinatory words that lack visual grounding. Our VA detector enables the detection of such visually unsupported tokens during generation. Specifically, after the model generates a token at iteration $t$, that token is subsequently fed back into the model as input at iteration $t+1$. The VA detector then leverages the FFN activations at iteration $t+1$ to determine if the previously generated token is visually absent. When the token is identified as visually absent, the generation process reverts back to the previous iteration, and the logit of the unsupported token is set to negative infinity, ensuring that the next most probable token is selected instead. This iterative rollback process continues at every decoding step, helping to suppress hallucination in free-form outputs.



\section{Experiments}\label{sec:exp}

\subsection{Experimental Settings}\label{sec:exp_setting}
\paragraph{Implementation Details}
We evaluate the effectiveness of our method on widely used LVLMs, including LLaVA variants (LLaVA-v1.5 (7B), LLaVA-v1.6 (7B), LLaVA-v1.6 (13B)) \citep{liu2024improved}, mPlug-Owl2 \citep{ye2024mplug}, InstructBLIP \citep{instructblip}, and recent models such as Qwen2-VL \citep{Qwen2-VL} and Gemma3 \citep{team2025gemma}. For each model, we train a detector using the VA-QA dataset and determine the optimal $\mathbf{S}^\text{VA}$ threshold $\beta$.  

\paragraph{Benchmark Datasets}
To verify the visual absence detection capability of our method, we use various hallucination evaluation benchmarks that assess whether models can correctly determine the presence or absence of specific objects or relations in an image. For binary question answering, we adopt our proposed VA-QA dataset, which is an in-domain dataset, as we trained the VA detector using the train split of the VA-QA dataset. For out-of-domain datasets, we evaluate our method on the widely used object and relation hallucination benchmarks, POPE \citep{li2023evaluating} and R-Bench \citep{wu2024evaluating}. To assess performance on a more general Visual Question Answering (VQA) setting, we employ SEED-Bench \cite{li2023seed}. Since our VA detector is particularly suitable for binary questions, we convert each multiple-choice question into multiple binary ones. For instance, a question such as ``Question: What is the color of the man’s suit? Options: A. Black B. White'' would be transformed into binary forms like ``Is the man’s suit black?'' and ``Is the man’s suit white?''. Finally, for open-ended generation, we use the CHAIR \citep{rohrbach2018object}.

\subsection{Experimental Results}\label{sec:exp_result}

\subsubsection{Binary Question Answering}

\begin{table*}[!t]
\centering
\caption{\textbf{Results on SEED-Bench.} All multiple-choice questions are converted to multiple binary questions. $\text{Acc}_{\text{yes}}$ and $\text{Acc}_{\text{no}}$ denote accuracies of questions with ground-truth answers `yes' and `no', respectively, and Acc represents the overall accuracy.}
\label{tab:app_seedb}
\begin{subtable}{\linewidth}
    \centering
    \resizebox{\linewidth}{!}{
    \begin{tabular}{llccccccccccccccc}
    \toprule
    \multirow{2}{*}{\textbf{Model}} &  &
    \multicolumn{3}{c}{\textbf{Instances Counting}} &
    \multicolumn{3}{c}{\textbf{Instance Attributes}} &
    \multicolumn{3}{c}{\textbf{Scene Understanding}} &
    \multicolumn{3}{c}{\textbf{Instance Identity}} &
    \multicolumn{3}{c}{\textbf{Instance Interaction}} \\
    \cmidrule(lr){3-5} \cmidrule(lr){6-8} \cmidrule(lr){9-11} \cmidrule(lr){12-14} \cmidrule(lr){15-17}
     & & $\text{Acc}_{\text{yes}}$ & $\text{Acc}_{\text{no}}$ & Acc & $\text{Acc}_{\text{yes}}$ & $\text{Acc}_{\text{no}}$ & Acc & $\text{Acc}_{\text{yes}}$ & $\text{Acc}_{\text{no}}$ & Acc & $\text{Acc}_{\text{yes}}$ & $\text{Acc}_{\text{no}}$ & Acc & $\text{Acc}_{\text{yes}}$ & $\text{Acc}_{\text{no}}$ & Acc \\
    \midrule
    \multirow{2}{*}{LLaVA-v1.5 (7B)} & Baseline & 95.6 & 31.3 & 47.5 & 96.4 & 37.5 & 52.2 & 96.1 & 46.2 & 58.6 & 90.7 & 36.3 & 49.9 & 94.8 & 29.2 & 45.5 \\
     & Ours     & 68.1 & 52.5 & \textbf{56.5} & 86.0 & 54.8 & \textbf{62.6} & 80.7 & 63.1 & \textbf{67.5} & 71.4 & 59.9 & \textbf{62.7} & 88.5 & 40.6 & \textbf{52.5} \\
    \midrule
    \multirow{2}{*}{mPLUG-Owl2 (7B)} & Baseline & 90.1 & 37.8 & 51.0 & 87.4 & 57.9 & 65.3 & 92.5 & 60.7 & 68.6 & 84.9 & 48.2 & 57.4 & 90.6 & 47.4 & 58.1 \\
     & Ours     & 65.9 & 59.4 & \textbf{61.1} & 56.2 & 78.9 & \textbf{73.2} & 67.5 & 77.4 & \textbf{74.9} & 58.0 & 76.4 & \textbf{71.8} & 68.8 & 67.4 & \textbf{67.7} \\
    \midrule
    \multirow{2}{*}{InstructBLIP (7B)} & Baseline & 87.7 & 27.0 & 42.3 & 62.7 & 79.8 & 75.5 & 85.4 & 64.9 & 70.0 & 73.6 & 60.7 & 63.9 & 71.9 & 65.6 & 67.2 \\
     & Ours     & 63.6 & 57.4 & \textbf{58.9} & 33.7 & 90.1 & \textbf{76.0} & 57.8 & 82.8 & \textbf{76.5} & 46.7 & 83.3 & \textbf{74.2} & 62.5 & 74.9 & \textbf{71.8} \\
    \midrule
    \multirow{2}{*}{Qwen2-VL (7B)} & Baseline & 87.9 & 48.6 & \textbf{58.5} & 89.6 & 71.1 & 75.7 & 91.5 & 66.8 & 72.9 & 79.5 & 64.3 & 68.1 & 85.4 & 54.3 & 62.0 \\
     & Ours     & 80.9 & 47.6 & 55.2 & 82.4 & 77.1 & \textbf{78.4} & 86.7 & 69.8 & \textbf{74.0} & 73.0 & 69.7 & \textbf{70.5} & 87.5 & 54.0 & \textbf{62.3} \\
    \bottomrule
    \end{tabular}
    }
\end{subtable}

\vspace{0.2cm}

\begin{subtable}{\linewidth}
    \centering
    \resizebox{0.85\linewidth}{!}{
    \begin{tabular}{llcccccccccccc}
    \toprule
    \multirow{2}{*}{\textbf{Model}} &  &
    \multicolumn{3}{c}{\textbf{Visual Reasoning}} &
    \multicolumn{3}{c}{\textbf{Instance Location}} &
    \multicolumn{3}{c}{\textbf{Spatial Relation}} &
    \multicolumn{3}{c}{\textbf{Text Understanding}} \\
    \cmidrule(lr){3-5} \cmidrule(lr){6-8} \cmidrule(lr){9-11} \cmidrule(lr){12-14}
     & & $\text{Acc}_{\text{yes}}$ & $\text{Acc}_{\text{no}}$ & Acc & $\text{Acc}_{\text{yes}}$ & $\text{Acc}_{\text{no}}$ & Acc & $\text{Acc}_{\text{yes}}$ & $\text{Acc}_{\text{no}}$ & Acc & $\text{Acc}_{\text{yes}}$ & $\text{Acc}_{\text{no}}$ & Acc \\
    \midrule
    \multirow{2}{*}{LLaVA-v1.5 (7B)} & Baseline & 93.4 & 45.5 & 57.5 & 97.8 & 18.2 & 38.1 & 95.3 & 14.0 & 34.3 & 97.6 & 22.3 & 41.2 \\
     & Ours     & 68.3 & 70.9 & \textbf{70.2} & 89.3 & 28.5 & \textbf{43.7} & 90.4 & 16.1 & \textbf{34.6} & 84.5 & 41.8 & \textbf{52.5} \\
    \midrule
    \multirow{2}{*}{mPLUG-Owl2 (7B)} & Baseline & 90.6 & 62.9 & 69.9 & 90.5 & 35.8 & 49.5 & 90.4 & 25.1 & 41.4 & 85.7 & 49.8 & 58.8 \\
     & Ours     & 63.1 & 83.8 & \textbf{78.6} & 65.1 & 59.4 & \textbf{60.8} & 66.1 & 46.5 & \textbf{51.4} & 31.0 & 80.9 & \textbf{68.4} \\
    \midrule
    \multirow{2}{*}{InstructBLIP (7B)} & Baseline & 77.0 & 70.6 & 72.2 & 69.2 & 51.8 & 56.2 & 74.5 & 37.7 & 46.9 & 19.0 & 91.2 & 73.1 \\
     & Ours     & 44.1 & 87.8 & \textbf{76.9} & 44.9 & 74.7 & \textbf{67.2} & 55.6 & 57.5 & \textbf{57.0} & 3.6 & 97.6 & \textbf{74.0} \\
    \midrule
    \multirow{2}{*}{Qwen2-VL (7B)} & Baseline & 82.5 & 71.5 & 74.2 & 89.9 & 47.0 & 57.7 & 90.4 & 33.1 & \textbf{47.4} & 85.7 & 65.9 & 70.8 \\
     & Ours     & 71.6 & 78.7 & \textbf{76.9} & 83.1 & 54.2 & \textbf{61.4} & 88.7 & 27.6 & 42.8 & 75.0 & 80.9 & \textbf{79.2} \\
    \bottomrule
    \end{tabular}
    }
\end{subtable}
\end{table*}

Our method leverages the internal recognition capability of VA neurons to correct model outputs by ensuring that questions with unsupported visual concepts are answered with ``No''. As shown in \cref{tab:exp_binary}, our approach consistently shows a significant improvement in $\text{Acc}_{\text{no}}$ across all models and datasets. For example, on the VA-QA dataset, it rises from 48.0\% to 77.5\% for LLaVA-v1.5, and also on R-Bench, it improves from 39.3\% to 51.7\% for the same model. These gains confirm that our refinement effectively enhances the models’ ability to correctly identify and reject misaligned inputs. While there is a slight trade-off with reduced $\text{Acc}_{\text{yes}}$ in some cases due to the model being more conservative, the overall accuracy ($\text{Acc}$) generally improves or remains comparable to the baseline. 

Furthermore, we evaluate our method on diverse question types in SEED-Bench, which serves as a more general VQA benchmark beyond hallucination-specific settings. \cref{tab:app_seedb} provides detailed evaluation results comparing the baseline and our method across each category. All models consistently outperform their baselines across every category, with Qwen-VL2 also showing improvements in all but two categories. These results demonstrate that our proposed method is robust and generalizable, even beyond the object- and relation-centric questions that were present in the training dataset.

\subsubsection{Open-ended Generation}
Our approach replaces detected visually absent tokens with more visually faithful alternatives during generation. To evaluate the extent of hallucination in model outputs, we employ CHAIR metrics. Specifically, for 500 randomly selected images from the MSCOCO dataset~\citep{lin2014microsoft}, we prompt LVLMs with ``Please describe this image in detail’’ to generate visual captions. CHAIR then quantifies hallucination by measuring the proportion of objects mentioned in the captions that are not present in the ground-truth object list, at both sentence-level ($C_s$) and instance-level ($C_i$):
\begin{align} 
    \text{C}_s &= \frac{\small\text{{\# of hallucinated objects}}}{\small\text{{\# of all objects mentioned}}}, \\[6pt] 
    \text{C}_i &= \frac{\small\text{{\# of sentences w/ hallucinative object}}}{\small\text{{\# of all sentences}}}.
\end{align}

As presented in \cref{tab:exp_chair_sum}, our method consistently reduces both $C_s$ and $C_i$ across most models, demonstrating its effectiveness in suppressing hallucinated content during generation. For instance, on LLaVA-v1.6, $C_s$ drops from 42.8 to 42.2 and $C_i$ from 23.5 to 23.0, and more pronounced improvements are seen in Gemma3, as $C_s$ reduced from 51.8 to 43.6 and $C_i$ from 23.5 to 22.2. In Qwen2-VL, a slight increase in $C_i$ occurs as the refinement shortens captions -- lowering the denominator -- despite a decrease in the absolute number of hallucinative objects. 

\begin{figure*}[!t]
    \centering
    \includegraphics[width=\linewidth]{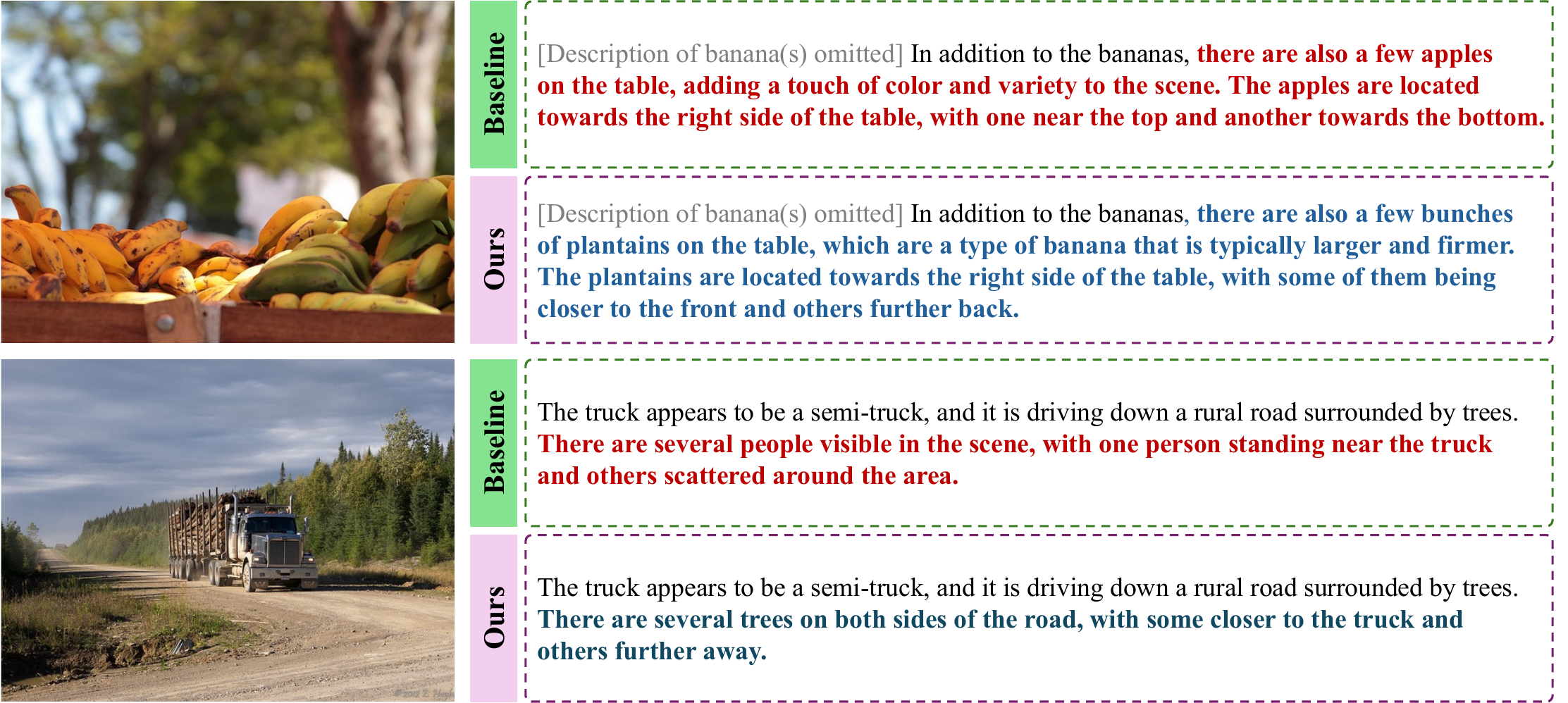}
    \caption{\textbf{Qualitative Results.} Comparison between a caption generated using LLaVA-1.5 and the refined version using our method.}
    \label{fig:exp_chair_qual}
\end{figure*}

\begin{table}[t]
\centering
\captionof{table}{\textbf{Results on Open-ended Generation.} `Length' indicates the generation length, and all results are extracted with max\_tokens = 300.}
\label{tab:exp_chair_sum}
\resizebox{\linewidth}{!}{
    \begin{tabular}{llcccc}
    \toprule
    \textbf{Model} & & $\text{C}_s$ $\downarrow$ & $\text{C}_i$ $\downarrow$ & Length & GPT-score $\downarrow$ \\ 
    \midrule
    \multirow{2}{*}{LLaVA-v1.5 (7B)} & Baseline & 60.0 & 29.0 & 100.3 & 111.6 \\ 
     & Ours & \textbf{58.6} & \textbf{28.1} & 100.2 & \textbf{111.3} \\
    \midrule
    \multirow{2}{*}{LLaVA-v1.6 (13B)} & Baseline & 42.8 & 23.5 & 182.1 & 123.1 \\ 
     & Ours & \textbf{42.2} & \textbf{23.0} & 178.7 & \textbf{115.2} \\ 
    \midrule
    \multirow{2}{*}{mPLUG-Owl2 (7B)} & Baseline & 66.8 & 30.6 & 105.1 & 130.7 \\ 
     & Ours& \textbf{57.2} & \textbf{27.9} & 104.2 & \textbf{116.6}\\ 
    \midrule
    \multirow{2}{*}{InstructBLIP (7B)} & Baseline & 65.2 & 37.3 & 108.4 & \textbf{201.5}\\ 
     & Ours & \textbf{49.4} & \textbf{30.8} & 106.4 & 206.0 \\ 
    \midrule
    \multirow{2}{*}{Qwen2-VL (7B)} & Baseline & 49.2 &\textbf{22.6} & 252.5 & 84.4 \\ 
     & Ours & \textbf{47.2} & 23.7 & 228.2 & \textbf{83.8} \\ 
    \midrule
    \multirow{2}{*}{Gemma3 (12B)} & Baseline & 51.8 & 23.5 & 294.5 & 71.7 \\ 
     & Ours & \textbf{43.6} & \textbf{22.2} & 293.3 & \textbf{62.6} \\ 
    \bottomrule
    \end{tabular}
}
\end{table}

Additionally, we conduct a GPT-based evaluation to assess the severity and frequency of hallucinated elements.\footnote{The prompt used for GPT-based evaluation is provided in the appendix.} The GPT-score also decreases across most of the models, confirming that our refinement strategy enhances factual grounding. Notably, these improvements are achieved with minimal impact on caption length, preserving the informativeness of the generated outputs. Also, as shown in \cref{fig:exp_chair_qual}, our method effectively removes hallucinative tokens and guides the model to generate more visually grounded responses.

\section{Conclusion}
\label{sec:conclusion}
We revealed that Large Vision-Language Models (LVLMs) often incorrectly interpret text inputs lacking visual grounding as visually supported, leading to erroneous responses. Through systematic analysis, we uncovered a specific subset of feed-forward network neurons -- Visual Absence-aware (VA) neurons -- that consistently exhibit distinct activation patterns in response to visually absent tokens. Building on this observation, we introduced a lightweight Visual Absence (VA) Detector capable of classifying such tokens based on VA neuron activations. We employed the VA detector to refine model responses by reevaluating question prompts and replacing detected tokens during generation, thereby aligning model outputs with its internal signals of visual absence. Experimental results across multiple models confirmed that our refinement strategy effectively mitigates hallucinations in both binary question answering and open-ended generation. Overall, our approach offered a promising avenue for improving factual grounding and reliability of LVLMs.

\section*{Limitations} 
While our method demonstrates promising results and broad applicability, a few limitations remain. First, it relies on internal neuron activations, which may not be directly accessible in closed-source LVLMs. Additionally, as the method does not incorporate external knowledge, it can only leverage information encoded during pre-training. However, its independence from model-specific fine-tuning makes it lightweight and broadly transferable across open-source models.

Second, the Visual Absence (VA) detector is trained on the VA-QA dataset, which primarily focuses on object- and relation-level grounding. Consequently, it is currently optimized to identify hallucinations at this level. Expanding the training data to include a wider variety of reasoning types -- such as attribute recognition or temporal understanding -- could enhance the detector's ability to identify and mitigate more diverse forms of hallucinations.

\section*{Acknowledgements} 
This work was supported by the Institute of Information \& Communications Technology Planning \& Evaluation (IITP) grant funded by the Korea government (MSIT) (No. RS-2024-00457882, AI Research Hub Project; No. RS-2025-02305581, Development of Vision-Language Model (VLM)-Based Intelligent Video Security Monitoring Technology; No. 2022-0-00713, Meta-learning applicable to real-world problems; No. RS-2025-II250075, Artificial Intelligence Graduate School Program(KAIST)). 
It was also supported by the National Research Foundation of Korea (NRF) grant (No. RS-2023-00209060, A Study on Optimization and Network Interpretation Method for Large-Scale Machine Learning) funded by the Korea government (MSIT).

\bibliography{main}

\clearpage
\appendix

\section{Additional Analysis of Visual Absence-aware Neurons in Various Models}

\begin{figure*}[b]
    \centering
    \includegraphics[width=0.85\linewidth]{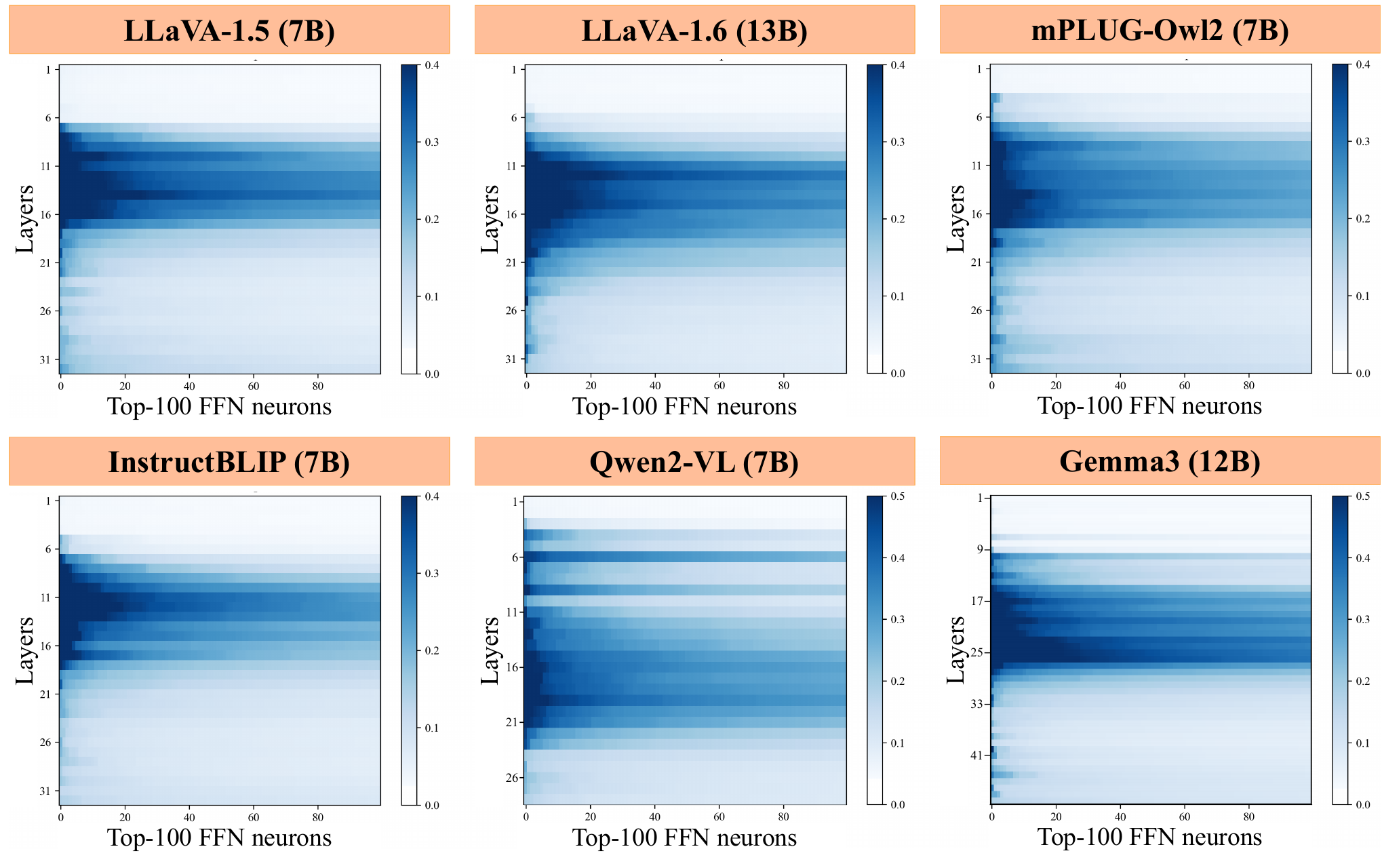}
    \caption{\textbf{Top-100 $\mathbf{S}^\text{VA}$ across layers.}}
    \label{fig:score_heatmap_all}
\end{figure*}

\begin{figure*}[t]
    \centering
    \includegraphics[width=\linewidth]{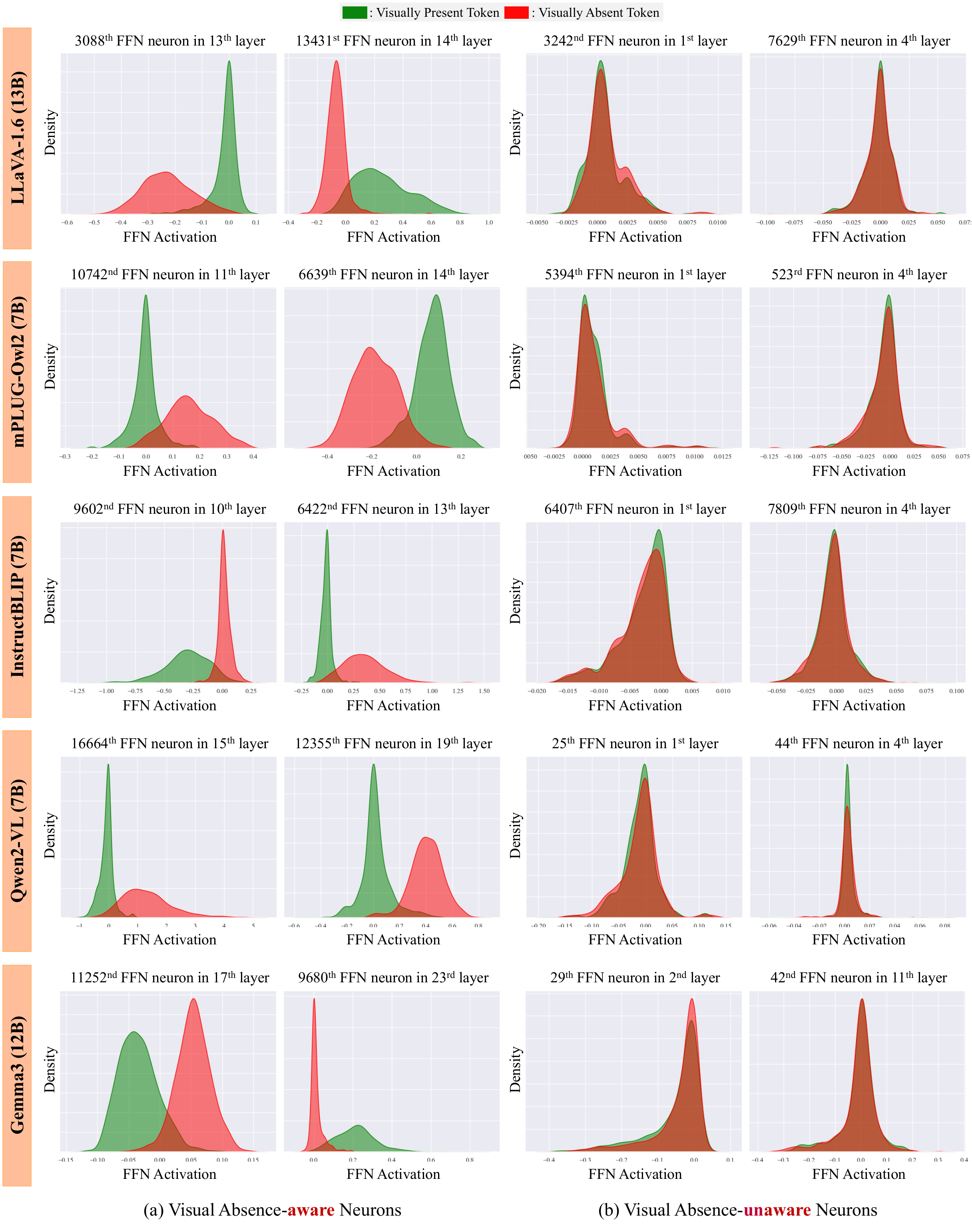}
    \caption{\textbf{FFN Activation Patterns in Response to Visual Absence.} This figure shows the distinct activation pattern of Visual Absence-aware neurons in response to visual absence. It demonstrates that all models possess a specific set of neurons that respond to visually absent tokens.}
    \label{fig:kde_all}
\end{figure*}

To assess the generality of our findings beyond LLaVA-v1.5, we extend the experiments from \cref{sec:emprical} to additional LVLMs which were examined in \cref{sec:exp}. The results, shown in \cref{fig:score_heatmap_all} and \ref{fig:kde_all}, demonstrate that the existence of Visual Absence-aware (VA) neurons is not confined to a single model. 

\cref{fig:kde_all} shows the activation distributions of FFN neurons in response to visually present and absent tokens. Across all models, we observe that specific neurons exhibit distinct activation shifts between the two conditions, highlighting their sensitivity to the absence of visual grounding.

Additionally, to investigate where these VA neurons are located within each model, we compute $\mathbf{S}^\text{VA}$ for all FFN neurons and visualize the top 100 values across layers in \cref{fig:score_heatmap_all}. The resulting heatmaps reveal that VA neurons are primarily concentrated in middle layers, consistent with the pattern observed in LLaVA-v1.5 (\cref{fig:score_heatmap}).

\clearpage
\section{Discrepancy between Visual Absence-aware Neuron Activations and Model Responses}
\begin{figure*}[t]
    \centering
    \includegraphics[width=\linewidth]{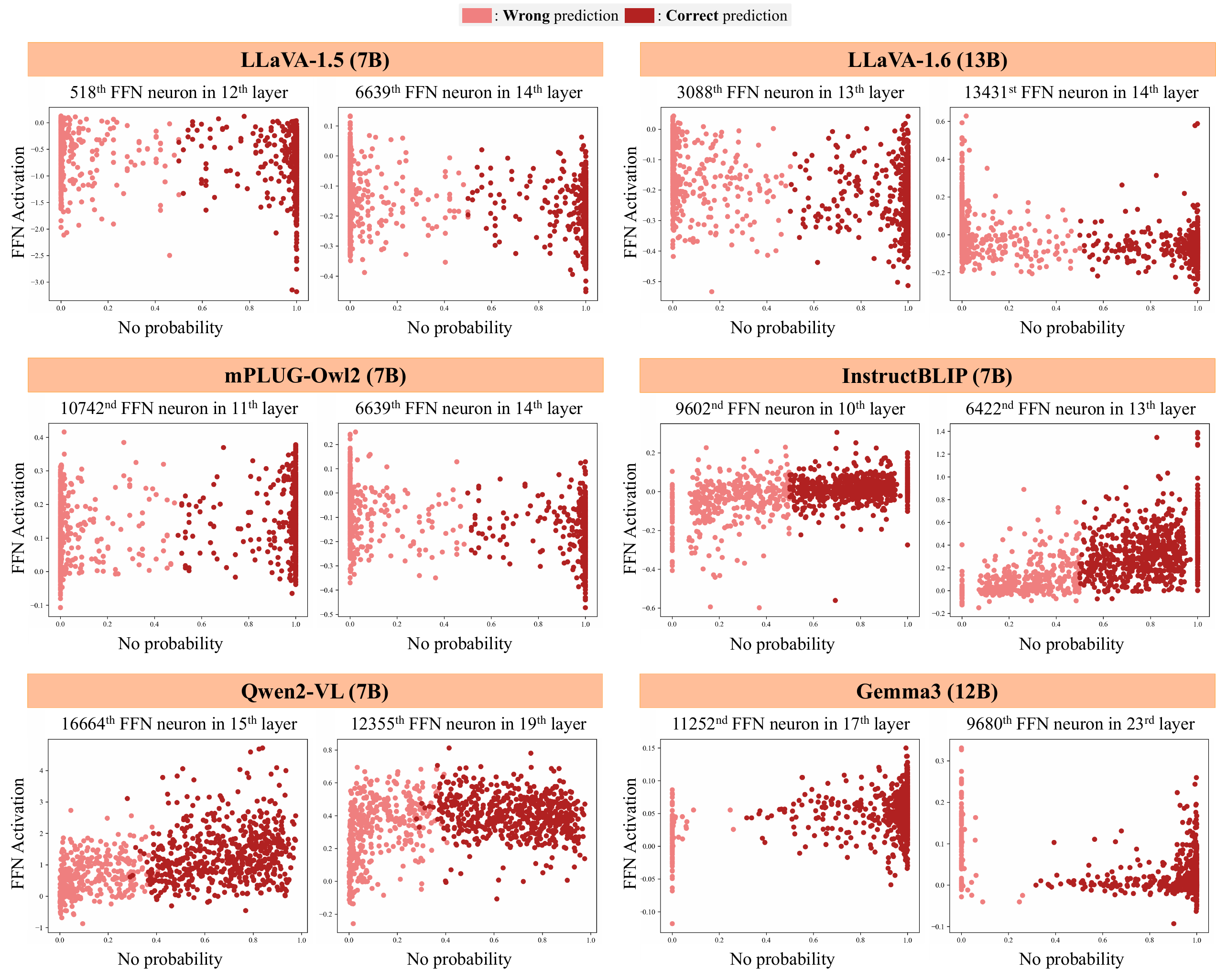}
    \caption{\textbf{VA Neuron Activation and No Probability.} For the VA-QA dataset, we visualize VA neuron activations for visually absent tokens in yes-or-no questions alongside the model's probability of predicting ``No.'' Ideally, the model should answer ``No'' due to the absence of relevant visual information, though it often fails. Nonetheless, regardless of whether the prediction is \textbf{\textcolor{Maroon}{correct}} or \textbf{\textcolor{Salmon}{wrong}}, VA neurons constantly exhibit high activation values.}
    \label{fig:prob_act}
\end{figure*}

\cref{fig:prob_act} presents the activation of Visual Absence (VA) neurons and the corresponding probability assigned to the answer ``No'' for questions in the VA-QA dataset across various LVLMs. We observe that VA neurons consistently exhibit high activation values when the input question contains visually absent tokens, indicating that the models are internally responsive to the absence of relevant visual information. However, the predicted probabilities for ``No'' remain low, indicating a disconnect between internal detection and output behavior. This discrepancy highlights the necessity for our proposed refinement method using the VA detector, which aims to align the model's response with its internal recognition of visual absence.

\clearpage
\begin{figure*}[t]
    \centering
    \includegraphics[width=\linewidth]{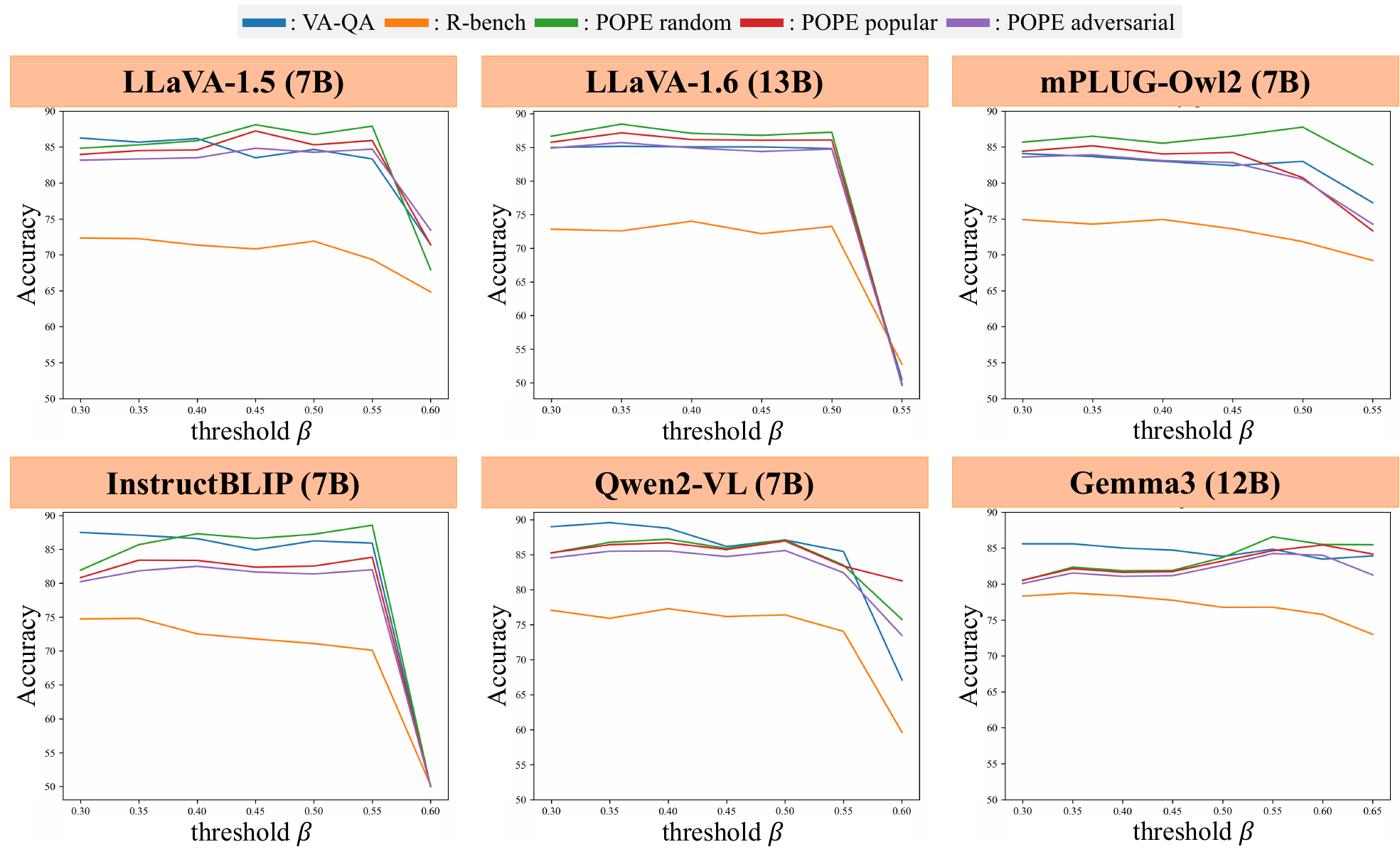}
    \caption{\textbf{Effect of Score Threshold $\beta$ Selection.} This figure shows the accuracy of binary question answering benchmarks across various models depending on the threshold $\beta$ used to select VA neurons.}
    \label{fig:threshold}
\end{figure*}

\section{Effect of Score Threshold Selection for VA Neuron Identification}
To examine the impact of the threshold hyperparameter $\beta$ used to identify Visual Absence-aware (VA) neurons, we conduct an ablation study across multiple LVLMs. As described in \cref{subsec:va_cls}, $\beta$ controls which neurons are selected based on their sensitivity score $\mathbf{S}^\textbf{VA}$, thereby determining the feature set used to train the VA detector. 

\cref{fig:threshold} illustrates the performance of the VA detector across different values of $\beta$, evaluated on various binary question answering datasets. We observe a consistent pattern across all models: accuracy peaks within a specific range of $\beta$, indicating that the neurons selected at an optimal threshold are most effectively reflecting the model's inherent capability to recognize visual absence. When $\beta$ is too low, the inclusion of less informative or noisy neurons dilutes the signal. Conversely, overly high thresholds result in too few neurons, weakening the feature representation. The detector trained using the optimal threshold demonstrates stable and high accuracy across all models, demonstrating that a well-chosen subset of VA neurons captures the model’s capacity for visual grounding most effectively.

\clearpage

\begin{table*}[!t]
\centering
\caption{\textbf{Results on SEED-Bench with VA Detector Trained with Diverse Data.}}
\label{tab:app_diverse_seedb}
\resizebox{0.7\linewidth}{!}{
\begin{tabular}{llcccccc}
\toprule
\multirow{2}{*}{\textbf{Model}} &  &
\multicolumn{3}{c}{\textbf{Instance Attributes}} &
\multicolumn{3}{c}{\textbf{Visual Reasoning}} \\
\cmidrule(lr){3-5} \cmidrule(lr){6-8}
 & & $\text{Acc}_{\text{yes}}$ & $\text{Acc}_{\text{no}}$ & Acc & $\text{Acc}_{\text{yes}}$ & $\text{Acc}_{\text{no}}$ & Acc \\
\midrule
\multirow{3}{*}{LLaVA-v1.5 (7B)} 
 & Baseline & 97.3 & 36.5 & 51.7 & 94.0 & 46.3 & 58.2 \\
 & Ours     & 87.5 & 53.8 & 62.3 & 67.2 & 68.2 & 67.9 \\
 & Ours w/ diverse data & 87.2 & 60.1 & \textbf{66.9} & 73.1 & 71.1 & \textbf{71.6} \\
\midrule
\multirow{3}{*}{mPLUG-Owl2 (7B)} 
 & Baseline & 89.4 & 58.4 & 66.2 & 92.6 & 60.2 & 68.3 \\
 & Ours     & 67.8 & 73.6 & 72.1 & 62.7 & 85.1 & 79.5 \\
 & Ours w/ diverse data & 56.9 & 79.7 & \textbf{74.0} & 65.7 & 86.1 & \textbf{81.0} \\
\midrule
\multirow{3}{*}{Qwen2-VL (7B)} 
 & Baseline & 90.4 & 71.8 & 76.5 & 80.6 & 69.2 & 72.0 \\
 & Ours     & 84.1 & 78.0 & 79.6 & 77.6 & 76.6 & 76.9 \\
 & Ours w/ diverse data & 74.4 & 83.6 & \textbf{81.3} & 73.1 & 78.6 & \textbf{77.2} \\
\bottomrule
\end{tabular}
}
\end{table*}

\begin{table*}[!t]
\centering
\caption{\textbf{Results on Binary Question Answering with VA Detector Trained with Diverse Data.}}
\label{tab:app_diverse_bin}
\resizebox{\linewidth}{!}{
\begin{tabular}{lccccccccccccccc}
\toprule
& \multicolumn{3}{c}{\textbf{In-domain}} & \multicolumn{12}{c}{\textbf{Out-of-domain}} \\ 
\cmidrule(lr){2-4} \cmidrule(lr){5-16}
& \multicolumn{3}{c}{\multirow{2}{*}{\textbf{VA-QA}}} & \multicolumn{3}{c}{\multirow{2}{*}{\textbf{R-Bench}}} & \multicolumn{9}{c}{\textbf{POPE}} \\
\cmidrule(lr){8-16}
& \multicolumn{3}{c}{} & \multicolumn{3}{c}{} & \multicolumn{3}{c}{Random} & \multicolumn{3}{c}{Popular} & \multicolumn{3}{c}{Adversarial} \\
\cmidrule(lr){2-4} \cmidrule(lr){5-7} \cmidrule(lr){8-10} \cmidrule(lr){11-13} \cmidrule(lr){14-16}
& $\text{Acc}_{\text{yes}}$ & $\text{Acc}_{\text{no}}$ & Acc 
   & $\text{Acc}_{\text{yes}}$ & $\text{Acc}_{\text{no}}$ & Acc 
   & $\text{Acc}_{\text{yes}}$ & $\text{Acc}_{\text{no}}$ & Acc 
   & $\text{Acc}_{\text{yes}}$ & $\text{Acc}_{\text{no}}$ & Acc 
   & $\text{Acc}_{\text{yes}}$ & $\text{Acc}_{\text{no}}$ & Acc \\
\midrule
Qwen2-VL (7B) & 87.3 & 72.7 & 80.0 & 87.9 & 62.5 & 75.3 & 84.6 & 97.5 & 90.9 & 84.6 & 92.9 & 88.8 & 84.6 & 89.1 & 86.8 \\
+ Ours     & 87.6 & 86.7 & 87.1 & 87.0 & 65.8 & 76.5 & 76.5 & 98.3 & 87.1 & 76.5 & 97.4 & 87.0 & 76.5 & 94.5 & 85.5 \\
+ Ours w/ diverse data & 92.2 & 80.8 & 86.5 & 92.7 & 58.6 & 75.8 & 81.3 & 97.2 & 89.0 & 81.3 & 94.9 & 88.1 & 81.3 & 91.9 & 86.6 \\
\bottomrule
\end{tabular}
}
\end{table*}

\section{Additional Results}
\subsection{Results on Additional Benchmark Dataset} \label{subsec:app_additional_benchmark}

\begin{table}[!h]
    \centering
    \caption{\textbf{Results on Winoground.}}
    \label{tab:app_winoground}
    \resizebox{\linewidth}{!}{
    \begin{tabular}{llccc}
    \toprule
    \textbf{Model} & & $\text{Acc}_{\text{yes}}$ & $\text{Acc}_{\text{no}}$ & Acc \\
    \midrule
    \multirow{2}{*}{LLaVA-v1.5 (7B)} & Baseline & 90.39 & 14.00 & 54.88 \\
     & Ours & 76.92 & 48.08 & \textbf{62.50} \\
    \midrule
    \multirow{2}{*}{mPLUG-Owl2 (7B)} & Baseline & 92.31 & 26.92 & 59.62 \\
     & Ours & 55.77 & 78.85 & \textbf{67.31} \\
    \midrule
    \multirow{2}{*}{InstructBLIP (7B)} & Baseline & 73.08 & 48.08 & 60.58 \\
     & Ours & 44.23 & 78.85 & \textbf{61.54} \\
    \midrule
    \multirow{2}{*}{Qwen2-VL (7B)} & Baseline & 76.92 & 51.92 & 64.42 \\
     & Ours & 75.00 & 55.77 & \textbf{65.38} \\
    \bottomrule
    \end{tabular}
    }
\end{table}

We evaluate our method on the Winoground dataset \cite{thrush2022winoground}, which contains paired images and captions. Similar to our constructed VA-QA dataset (\cref{subsec:emp_prob_def}), each pair of captions in Winoground consists of identical words arranged in different orders. To apply our approach, we convert each caption into a binary question, resulting in two “yes” questions and two “no” questions per pair. We focus on the \textit{Both} split of Winoground, as it contains captions with visually absent words, providing a challenging setting to assess our method. As shown in \cref{tab:app_winoground}, our approach achieves consistent improvements under these conditions.

\subsection{Training VA detector with Diverse Data}

As we acknowledged in the Limitations section, our VA detector primarily targets object- and relation-level grounding, and we anticipated potential weaknesses when applied to a wider variety of question types. To examine this aspect empirically, we conducted evaluations using two specific categories from SEED-Bench \cite{li2023seed}: ``Instance Attributes'' for attribute-based hallucinations and ``Visual Reasoning'' for assessing visual reasoning capabilities. Since SEED-Bench questions are multiple-choice-based, and our method is applicable to binary (yes-or-no) questions, we transformed each multiple-choice question into a series of binary questions as explained in \cref{sec:exp}.

Interestingly, as illustrated in \cref{tab:app_diverse_seedb}, our method effectively handles both attribute-based hallucinations and reasoning-based questions, thereby demonstrating that its generalizability extends beyond the scope we initially expected. Moreover, as mentioned in our Limitations section, we assumed that further training the VA detector on data covering diverse hallucination types would lead to even greater effectiveness. To support this claim experimentally, we train our VA detector on data from each SEED-Bench category mentioned above (with an 8:2 train-test split). The improved results on the test split confirm the benefit of incorporating a wider variety of hallucination types into the training set.
 
Additionally, a VA detector trained on diverse data also enhances performance on binary benchmark datasets shown in \cref{tab:exp_binary}. Since Qwen2-VL exhibited the most unstable performance, weevaluatede its performance. As shown in \cref{tab:app_diverse_bin}, enriching the training data improved the performance of the VA detector compared to our previous results. Given the clear improvement from adding even a single category of questions, we anticipate that training with a broader and more diverse dataset would yield a more robust and generalizable VA detector across various scenarios.

\subsection{Generation with Various Decoding Strategies}

\begin{table}[!h]
\centering
\caption{\textbf{Results on Open-ended Generation with Beam Search Decoding.}}
\label{tab:app_beam}
\resizebox{\linewidth}{!}{
\begin{tabular}{llcccc}
\toprule
\textbf{Model} & & $\text{C}_s$ $\downarrow$ & $\text{C}_i$ $\downarrow$ & Length & GPT-score $\downarrow$ \\ 
\midrule
\multirow{2}{*}{LLaVA-v1.5 (7B)} 
 & Baseline & 63.8 & 30.0 & 104.5 & 118.1 \\
 & Ours     & \textbf{62.8} & \textbf{29.5} & 105.4 & \textbf{91.8} \\
\midrule
\multirow{2}{*}{mPLUG-Owl2 (7B)} 
 & Baseline & 67.2 & 31.8 & 105.8 & 135.2 \\
 & Ours     & \textbf{60.8} & \textbf{29.1} & 110.8 & \textbf{124.7} \\
\midrule
\multirow{2}{*}{InstructBLIP (7B)} 
 & Baseline & 57.4 & 28.7 & 99.3  & 124.6 \\
 & Ours     & \textbf{36.6} & \textbf{22.4} & 116.6 & \textbf{104.0} \\
\midrule
\multirow{2}{*}{Qwen2-VL (7B)} 
 & Baseline & 52.0 & 23.1 & 253.2 & 75.6 \\
 & Ours     & \textbf{51.4} & \textbf{23.0} & 254.9 & \textbf{73.6} \\
\bottomrule
\end{tabular}
}
\end{table}

Our refinement method is independent of decoding strategies as it solely adjusts logit probabilities. It is thus compatible with diverse decoding strategies beyond greedy decoding, including sampling strategies (e.g., setting temperature, top-p, top-k) as well as beam search.

Specifically, in beam search decoding, our method can be directly applied by evaluating each candidate beam individually through the VA detector. When visually absent tokens are detected, the process rolls back to previous tokens within each candidate beam and applies a scoring penalty accordingly. \cref{tab:app_beam} demonstrates consistent performance improvements obtained by applying our method in combination with beam search. These results confirm that our approach is robust and decoding strategy-agnostic.

\subsection{Combination with Existing Hallucination-Mitigation Strategies}

\begin{table}[!t]
\centering
\caption{\textbf{Results on Open-ended Generation with Existing Hallucination-Mitigation Strategies.}}
\label{tab:app_dec_comb}
\resizebox{\linewidth}{!}{
\begin{tabular}{llcccc}
\toprule
\textbf{Model} & & $\text{C}_s$ $\downarrow$ & $\text{C}_i$ $\downarrow$ & Length & GPT-score $\downarrow$ \\ 
\midrule
\multirow{6}{*}{LLaVA-v1.5 (7B)} 
 & Greedy        & 60.0 & 29.0 & 100.3 & 111.6 \\
 & Greedy + Ours & \textbf{58.6} & \textbf{28.1} & 100.2 & \textbf{111.3} \\
 & VCD           & 58.2 & 28.9 & 100.6 & 113.8 \\
 & VCD + Ours    & \textbf{57.6} & \textbf{28.6} & 100.4 & \textbf{110.7} \\
 & DoLA          & 59.0 & 28.9 & 99.0  & 116.9 \\
 & DoLA + Ours   & \textbf{57.0} & \textbf{28.4} & 98.4  & \textbf{115.9} \\
\midrule
\multirow{6}{*}{mPLUG-Owl2 (7B)} 
 & Greedy        & 66.8 & 30.6 & 105.1 & 130.7 \\
 & Greedy + Ours & \textbf{57.2} & \textbf{27.9} & 104.2 & \textbf{116.6} \\
 & VCD           & 68.0 & 31.7 & 105.6 & 134.3 \\
 & VCD + Ours    & \textbf{50.4} & \textbf{27.2} & 106.9 & \textbf{111.3} \\
 & DoLA          & 67.4 & 30.8 & 104.6 & 136.1 \\
 & DoLA + Ours   & \textbf{48.1} & \textbf{29.9} & 91.6  & \textbf{82.0} \\
\midrule
\multirow{6}{*}{Qwen2-VL (7B)} 
 & Greedy        & 49.2 & \textbf{22.6} & 252.5 & 84.4 \\
 & Greedy + Ours & \textbf{47.2} & 23.7 & 228.2 & \textbf{83.8} \\
 & VCD           & 54.2 & \textbf{24.1} & 257.4 & 84.3 \\
 & VCD + Ours    & \textbf{50.4} & 24.4 & 233.6 & \textbf{79.7} \\
 & DoLA          & 43.2 & \textbf{22.0} & 230.0 & 79.1 \\
 & DoLA + Ours   & \textbf{42.6} & 23.1 & 209.5 & \textbf{73.1} \\
\bottomrule
\end{tabular}
}
\end{table}

Our method specifically aims at maximizing the utility of internally pre-trained knowledge within LVLMs, without relying on external sources or post-training techniques. Our experimental comparisons thus focused primarily on demonstrating improvements in performance relative to their raw outputs, quantifying how much better we can utilize the model’s internal knowledge through our approach. Because our method identifies visually absent tokens based on FFN activations of generated tokens during the decoding phase, it can be combined seamlessly with various other hallucination-mitigation strategies, as long as the underlying model terminology remains consistent. To illustrate this point, we present results combining our approach with two representative decoding-based methods, VCD \cite{leng2024mitigating} and DoLA \cite{chuang2023dola}.

For example, when integrated with VCD, we maintain VCD’s logit probability calculation procedure; however, when a generated token re-enters as input and is detected as visually absent by our VA detector, the decoding process rolls back one iteration and adjusts the logit probabilities accordingly, prompting the model to generate an alternative token. A detailed description of our open-ended generation refinement method is provided in \cref{subsec:va_decode} and \cref{subsec:app_gen_detail}.

\cref{tab:app_dec_comb} demonstrates consistent improvements achieved when integrating our method with VCD and DoLA, confirming its compatibility and effectiveness with existing training-free hallucination mitigation strategies.

\subsection{Evaluation on High-Capacity LVLMs}

\begin{table}[!t]
\centering
\caption{\textbf{Results on Open-ended Generation of Qwen2.5-VL}}
\label{tab:app_qwen25}
\resizebox{\linewidth}{!}{
\begin{tabular}{llcccc}
\toprule
\textbf{Model} & & $\text{C}_s$ $\downarrow$ & $\text{C}_i$ $\downarrow$ & Length & GPT-score $\downarrow$ \\ 
\midrule
\multirow{2}{*}{Qwen2.5-VL (7B)} 
 & Baseline & 44.8 & 23.7 & 176.1 & 60.0 \\
 & Ours     & \textbf{35.0} & \textbf{23.2} & 136.0 & \textbf{59.8} \\
\midrule
\multirow{2}{*}{Qwen2.5-VL (32B)} 
 & Baseline & 56.8 & 24.9 & 298.9 & 78.8 \\
 & Ours     & \textbf{47.2} & \textbf{24.1} & 244.3 & \textbf{62.0} \\
\bottomrule
\end{tabular}
}
\end{table}

\cref{tab:app_qwen25} shows clear performance improvements on Qwen2.5-VL \cite{bai2025qwen2} for both model sizes (7B and 32B), reaffirming the effectiveness of our method on powerful, high-capacity LVLMs and demonstrating the continued relevance of addressing vulnerabilities to visually absent tokens.

\clearpage
\section{Open-ended Generation Qualitative Results}
\begin{figure*}[b]
    \centering
    \includegraphics[width=\linewidth]{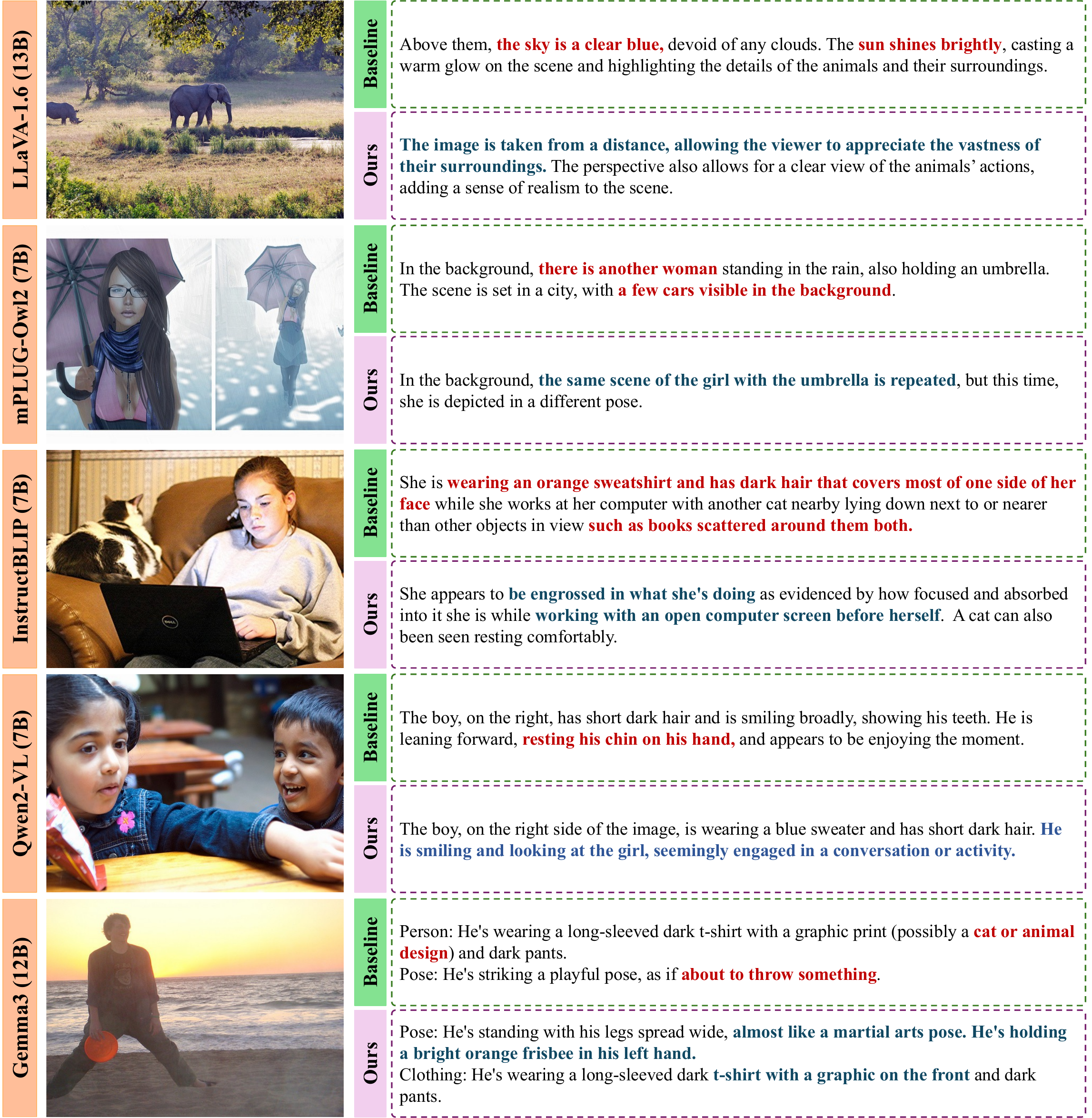}
    \caption{\textbf{Open-ended Generation Qualitative Results.}}
    \label{fig:gen_qual1}
\end{figure*}

We show the qualitative results of our generation refinement method across various models in \cref{fig:gen_qual1}, excluding the LLaVA-1.5 already presented in the main text. The red colored texts are visually ungrounded words, and the blue colored texts are the words that were revised by our method. As shown in these figures, our method effectively detects visually absent tokens and refines them to words that align with the given image.

\clearpage
\section{Experimental Details}

\subsection{Details of VA Detector-based Refinement for Open-ended Generation}\label{subsec:app_gen_detail}
We observe that VA neuron activations tend to weaken as the length of generated output increases, which can degrade the model's ability to detect visual absence in open-ended generation tasks. To address this, we mask previously generated sentences except the most recent one when extracting activations. Additionally, if the model repeatedly generates a visually absent token at the same position, we gradually roll back towards earlier tokens until it produces a more accurate and visually grounded response. Specifically, we progressively deepen the rollback by an additional step each time two rollbacks occur within a 5-token window. All experiments presented in the paper employ greedy text decoding to ensure determinism and eliminate randomness. 

\subsection{Implementation Details and Fit Quality of VA Detector}
To construct the Visual Absence-aware Detector, we utilized sklearn~\citep{scikit-learn} packages to train an MLPClassifier with a single hidden layer of 128 units. This detector was trained on features derived from VA neurons. From the original VA-QA dataset of 2400 samples, we first curated a subset consisting exclusively of instances answered correctly by each model individually. This curated subset was then partitioned into training and validation sets using a 9:1 ratio. To identify the optimal threshold hyperparameter $\beta$, which is used to select VA neurons, we search values from 0.3 to 0.8 in increments of 0.05, selecting $\beta$ that maximizes the accuracy on the validation set. For all evaluations on downstream tasks, we employed greedy decoding to ensure deterministic outputs.

To quantitatively assess the performance of our VA detector, \cref{tab:fit_quality_va} illustrates the fit quality of our VA detector for each LVLM, including metrics such as precision, recall, and accuracy. Despite being trained on a small amount of data with a lightweight 2-layer MLP classifier, the results demonstrate strong fitting quality across all metrics. This confirms that the activation patterns of VA neurons exhibit consistency and robustness across different models.

\begin{table}[t]
\centering
\small
\caption{\textbf{Fit Quality of VA detector.}}
\label{tab:fit_quality_va}
\begin{tabular}{lccc}
\toprule
\textbf{Model} & \textbf{Precision} & \textbf{Recall} & \textbf{Accuracy} \\
\midrule
LLaVA-v1.5 (7B)     & 0.965 & 0.943 & 97.6 \\
LLaVA-v1.6 (13B)    & 0.940 & 0.979 & 97.6 \\
mPLUG-Owl2 (7B)     & 0.938 & 0.958 & 97.0 \\
InstructBLIP (7B)   & 0.980 & 0.978 & 98.8 \\
Qwen2-VL (7B)       & 0.951 & 0.934 & 97.2 \\
Gemma3 (12B)        & 0.993 & 0.929 & 97.4 \\
\bottomrule
\end{tabular}
\end{table}

\subsection{Robustness of VA Detector}
When constructing the dataset used to train our VA detector, we explicitly excluded ambiguous cases, selecting only clearly and definitively visually absent concepts (\cref{fig:dataset}). Thus, our VA detector is specifically tuned to detect only tokens that are unambiguously absent from the visual content, avoiding detection of ambiguous or partially grounded tokens. 

Furthermore, there might be cases where the questions contain visually absent tokens yet require a ``Yes'' answer, so we constructed an additional evaluation dataset specifically designed for these cases. This dataset covers role-based questions: these questions inquire about the role or purpose of objects in the image. For instance, if an image shows a hanger without any clothes and the question asks, ``Is this object used for hanging clothes?''---though ``clothes'' is visually absent, the correct answer remains ``Yes.''

As demonstrated in \cref{tab:absent_but_yes}, our method achieves strong performance in these scenarios as well, highlighting its robustness to such nuanced cases.

\begin{table}[t]
\centering
\caption{\textbf{Accuracy on Role-based Questions.}}
\label{tab:absent_but_yes}
\resizebox{0.5\linewidth}{!}{
\begin{tabular}{lc}
\toprule
\textbf{Model} & \textbf{Acc} \\
\midrule
LLaVA-v1.5 (7B)   & 90.0 \\
mPLUG-Owl2 (7B)   & 98.0 \\
Qwen2-VL (7B)     & 96.0 \\
\bottomrule
\end{tabular}
}
\end{table}

\subsection{Details of Evaluation Datasets}

\paragraph{\textbf{VA-QA (our constructed dataset)}} Based on SVO-probe, we construct a set of yes-or-no questions for each image using the corresponding \textlangle subject, verb, object\textrangle \ triplet. As detailed in \cref{subsec:emp_prob_def} : (1) \ Positive questions (answer=Yes) directly utilize the provided triplet, accurately describing the image. (2) Negative questions (answer=No) are formed by altering one element of the original triplet to create a mismatch with the image, thereby differing only in that single modification from the ``Yes'' question.

\paragraph{\textbf{R-bench~\citep{wu2024evaluating}}}
Unlike POPE, which only targets object hallucination, R-bench is a dataset for assessing hallucinations related to inter-object relationships. Based on MSCOCO~\citep{lin2014microsoft} captions, they extract relationship triplets by applying a scene graph parser and convert them into a yes-or-no question. The negative question (answer = No) is constructed by varying one element from the triplet. We only use the balanced subset, which has an equal number of positive and negative questions, for unbiased evaluation.

\paragraph{\textbf{POPE~\citep{li2023evaluating}}} The Polling-based Object Probing Evaluation (POPE) dataset is designed to assess 'object' hallucination in LVLMs. It evaluates the model's ability to answer yes-or-no questions in the form: ``Is there a \textlangle object\textrangle 
 in the image?''. For objects detected in the image, questions are constructed with the answer ``Yes''. Conversely, questions with the answer ``No'' are divided into 3 variants-\textit{Random}, \textit{Popular}, and \textit{Adversarial}- based on the frequency and co-occurrence of the absent objects.

\paragraph{\textbf{CHAIR~\citep{rohrbach2018object}}}
The Caption Hallucination Assessment with Image Relevance (CHAIR) metric is designed to quantify object hallucination in generated image captions. It operates by comparing object words mentioned in a caption against the ground truth object annotations of the corresponding image. Specifically, CHAIR measures the proportion of hallucinated object words, which are defined as object words in the caption that do not appear in the image's ground truth annotation list. A lower CHAIR score indicates less hallucination and thus better caption quality in terms of object grounding.

\subsection{GPT-4o assisted Open-ended Generation Evaluation}
Even though CHAIR is the most widely used metric for evaluating object existence, it has limitations: it does not cover all objects in the image and fails to address other types of hallucination, such as those related to attributes or relations. Therefore, we employ GPT-4o~\cite{hurst2024gpt}, a leading multi-modal model, to assist in our generation evaluation. As shown in \cref{tab:prompt_template}, we design a prompt that mimics the CHAIR evaluation process. This prompt guides the model to identify both visually grounded and hallucinative elements within the generated captions, while also scaling the severity of each identified hallucination. For each data, we extract a list of severity scores for the hallucinative elements, and calculate an individual hallucination score $s_{\text{hall}}$ for each data by summing the generated list of severity scores. Then we determine the final reported metric by taking the mean of these values across all data and multiplying it by 10.
\begin{table*}[htb!]
    \centering 
    \begin{minipage}{\linewidth}
        \noindent\rule{\linewidth}{0.4pt} 
        \vspace{0.8em} 
        {\ttfamily 
        \noindent\textless{}ASSISTANT\textgreater{} \{caption generated by the model\}
        
        \noindent\textless{}INSTRUCTION\textgreater{} We would like to request your feedback on the performance of the AI assistantino image caption generation of the given image.
        \par\addvspace{0.5\baselineskip} 

        Please evaluate the \textless{}ASSISTANT\textgreater{}'s caption against the image by listing the elements in two bullet-point categories:
        \par\addvspace{0.5\baselineskip}

        Correct Elements: a list of phrases in the caption that are directly and objectively grounded in the image.
        \par\addvspace{0.5\baselineskip}

        Hallucinated Elements: a list of phrases that describe visual content that does not appear in the image. For each hallucinated phrase, add a severity score from 1 to 5 in parentheses.
        \par\addvspace{0.2\baselineskip} 

        \hspace*{1em}Use the following scale:\par 
        \hspace*{2em}- 1 (Minimal): Slight over-description or plausible extrapolation; not directly visible but weakly implied.\par
        \hspace*{2em}- 2 (Low): Mild misstatement; plausible but not quite accurate.\par
        \hspace*{2em}- 3 (Moderate): Clearly ungrounded; some viewers may misinterpret presence based on image context.\par
        \hspace*{2em}- 4 (High): Definitely not in the image, and no contextual cues suggest it.\par
        \hspace*{2em}- 5 (Severe): Fabricated object or scene element with strong visual contradiction --- impossible or clearly absent.\par 
        \addvspace{0.5\baselineskip}

        Output correct and hallucinated elements in a **flat list format**([]), separately (do not output in bullets). 
        Do not add additional explanation.
        \par 
        } 

        \noindent\rule{\linewidth}{0.4pt} 

    \end{minipage}
    \caption{\textbf{Prompt template for GPT-4o for generation evaluation.} We use the GPT-4o's ability to evaluate the quality of the generation, by instructing it to generate correct elements and hallucinated elements in the caption generated by the model. We further use this result to make a hallucination score.}
    \label{tab:prompt_template}
\end{table*}

\end{document}